%% file: main.tex
\documentclass{article}

\usepackage{microtype}
\usepackage{graphicx}
\usepackage{subfigure}
\usepackage{dblfloatfix}
\usepackage{pifont}
\usepackage{booktabs} 
\usepackage{enumitem}
\usepackage{bbm}
\usepackage[table,dvipsnames]{xcolor}
\usepackage{colortbl}
\usepackage{multicol}
\usepackage{hyperref}

\usepackage{hyperref}

\newcommand{\xxnote}[3]{}
\renewcommand{\xxnote}[3]{\color{#2}{#1: #3}}

\newcommand{\diff}[1]{{\texttt{\small{diff}}}}

\newcommand{\method}{\diff{} history}

\newcommand{\cmarkb}{\textcolor[RGB]{0, 0, 0}{\ding{51}}}
\newcommand{\xmarkb}{\textcolor[RGB]{0, 0, 0}{\ding{55}}}

\newcommand{\dataset}{\texttt{\small{LangHack}}}

\definecolor{diffgreen}{RGB}{0,153,102}
\definecolor{diffgreen_scaled}{RGB}{30, 179, 70}

\usepackage[accepted]{icml2024}

\usepackage{amsmath}
\usepackage{amssymb}
\usepackage{mathtools}
\usepackage{amsthm}

\usepackage[capitalize,noabbrev]{cleveref}

\theoremstyle{plain}

\theoremstyle{definition}

\theoremstyle{remark}

\usepackage[textsize=tiny]{todonotes}

\icmltitlerunning{\texttt{diff} History for Neural Language Agents}

\begin{document}

\twocolumn[
\icmltitle{\texttt{diff} History for Neural Language Agents}




\begin{icmlauthorlist}
\icmlauthor{Ulyana Piterbarg}{yyy}{}\hspace{0.2in}
\icmlauthor{Lerrel Pinto}{yyy}{} \hspace{0.2in}
\icmlauthor{Rob Fergus}{yyy}{}\\
\end{icmlauthorlist}

\icmlaffiliation{yyy}{Dept of Computer Science, Courant Institute, NYU}

\icmlcorrespondingauthor{Ulyana Piterbarg}{up2021@cims.nyu.edu}

\icmlkeywords{Foundation Models, Decision-Making, Imitation Learning, Large Language Models, Vision-Language Models}

\vskip 0.3in
]


\printAffiliationsAndNotice{}  

\input{abstract}
\input{introduction}

\input{related_work}

\input{method}

\input{results}

\input{discussion}

\bibliography{main}
\bibliographystyle{icml2024}

\newpage
\input{appendix}


\end{document}

%% file: abstract.tex
\begin{abstract}

Neural Language Models (LMs) offer an exciting solution for general-purpose embodied control. However, a key technical issue arises when using an LM-based controller: environment observations must be converted to text, which coupled with history, results in long and verbose textual prompts. As a result, prior work in LM agents is limited to restricted domains with small observation size as well as minimal needs for interaction history or instruction finetuning. In this paper, we introduce \method{}, a simple and highly effective solution to these issues. By applying the Unix \diff{} command on consecutive text observations in the interaction histories used to prompt LM policies, we can both abstract away redundant information and focus the content of textual inputs on  the salient changes in the environment. On NetHack, an unsolved video game that requires long-horizon reasoning for decision-making, LMs tuned with \method{} match state-of-the-art performance for neural agents while needing $1800\times$ fewer training examples compared to prior work. Even on the simpler BabyAI-Text environment with concise text observations, we find that although \method{} increases the length of prompts, the representation it provides offers a 25\% improvement in the efficiency of low-sample instruction finetuning. Further, we show that \method{} scales favorably across different finetuning dataset sizes. We open-source our code and data to \texttt{\small{\url{https://diffhistory.github.io}}}.
\end{abstract}

%% file: introduction.tex
\section{Introduction}

Scale is a key driver of performance in classical supervised learning \cite{kaplan2020scaling}. However, when it comes to supervised learning for sequential decision-making, the picture is more murky: in certain domains, model improvement has been shown to scale log-linearly and sub log-linearly with task-specific demonstration count \cite{tuyls2023scaling, piterbarg2023nethack}. Sequential decision-making data is cumbersome and expensive to collect for most domains, especially for real-world problems, rendering sheer data scaling an untenable solution to improving neural agents.

From the cognitive science literature, we know that humans and animals leverage abstraction to rapidly master new behaviors from limited demonstrations and interactions \cite{gopnik2004theory, tenenbaum2011grow, allen2020rapid}. Many recent works including ``Scratchpad,'' ``Chain-of-Thought,'' and ``Tree-of-Thought'' have looked to abstraction as a means of improving the downstream generalization of neural language models (LMs) on complex tasks \cite{nye2021show, wei2022chain, yao2023tree, besta2023graph, wu2023spring}. However, the abstractions behind these works are task-specific. 

We propose \method{}, a method that introduces a highly effective and \textit{task-agnostic} abstraction into LM prompts in order to improve the quality of text generation for  sequential decision-making. When finetuning LMs on demonstrations of interaction sequences consisting of interleaved environment observations and expert actions, we substitute \textit{full-text} natural language observations for symbolically annotated but human-interpretable \textit{delta} observations. We refer to such an interaction demonstration sequence as a \method{}. Computing a \method{} is easy: just run the Unix command \texttt{\small{diff}} \cite{kernighan1984unix} between consecutive pairs of past full-text observations. During inference, we iteratively prompt LMs to generate natural language actions conditioned on recent \method{} at each timestep of environment interaction.

\begin{figure*}[ht!]
    \centering
    \includegraphics[width=0.93\textwidth]{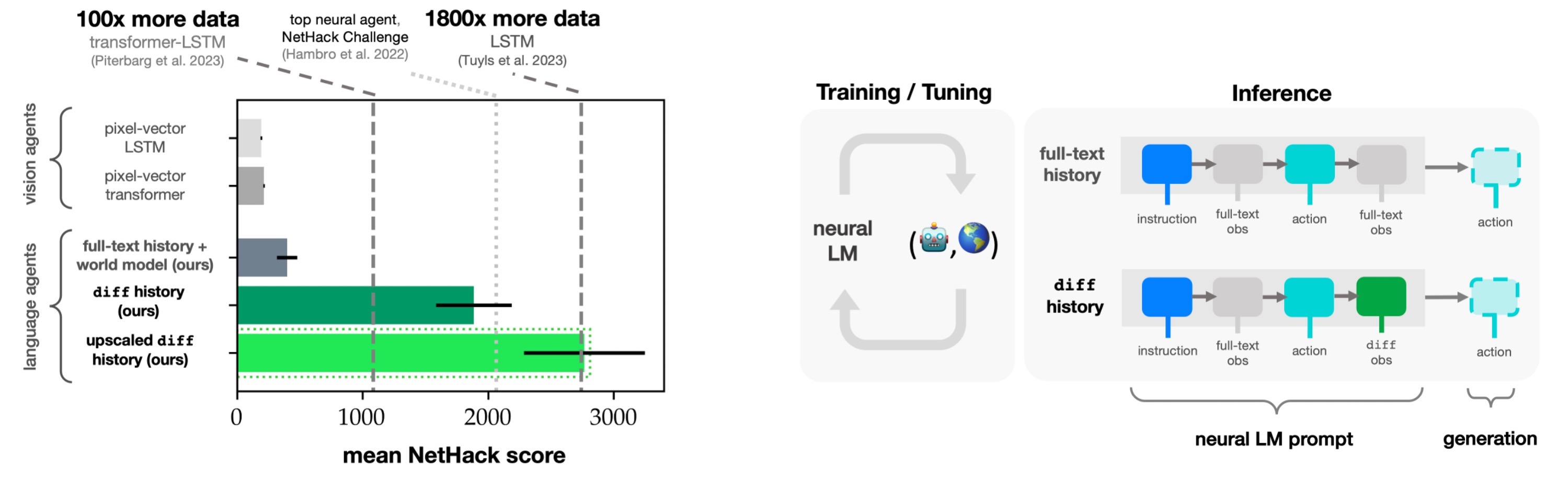}
    \caption{\textbf{Instruction finetuning LMs for sequential decision-making with \method{}}: We summarize our key results on the unsolved video-game NetHack (left) and illustrate our proposed approach (right). Neural language agents with \method{} match state-of-the-art performance on NetHack while training on orders of magnitude fewer data. \method{} improves the quality of generations in low-resource decision-making domains by replacing full-text observations in interaction history data provided in-context to LMs as ``memory'' with richly structured \textit{deltas}. These deltas are computed using an off-the-shelf implementation of the Unix \diff{} operation.}
    \vspace{-2mm}
    \label{fig:intro_figure}
\end{figure*}

We conduct a series of experiments testing instruction finetuning with \method{} in two different environments: the low-dimensional, multi-task BabyAI-Text environment \cite{carta2023grounding} and the high-dimensional, as-of-yet unsolved video game NetHack \cite{kuttler2020nethack}. Both of these environments are procedurally-generated, enabling us to directly probe the generalization of neural language agents in our study.  We find that \method{} has several benefits:

\begin{enumerate}
    \vspace{-2mm}
     \item It summarizes relational changes between text observations at consecutive timesteps of interaction, providing a dense learning signal for finetuning (Figure \ref{fig:diff_visualization}, Section \ref{section:df_hist_computing}, Appendix \ref{appendix:interaction_history_examples}).
    \item It can function as a ``soft compression'' mechanism for interaction histories in high-dimensional environments, providing an increase in the memory horizon available to language agents (Section \ref{section:diff_hist_whys}, Section \ref{subsection:diff_compresion}).
    \item In BabyAI-Text, neural language agents with full-text history require 33\% more gradient updates to match the performance of \method{} agents (Section \ref{section:results_action_pred}).
    \item  In NetHack, neural language agents with  \method{} attain state-of-the-art neural agent performance despite finetuning on $1800\times$ fewer data (Figure \ref{fig:intro_figure}). Ablating \diff{} observations from finetuning reduces neural language agent score on NetHack by 98\%, which suggests that \method{} is responsible for the highly efficient representation learning that we observe (Section \ref{section:results_action_pred}). The performance of \method{} agents scales with data (Section \ref{section:results_scaling}).
\end{enumerate}

%% file: related_work.tex
\section{Related Work}
\label{section:related_work}

\paragraph{Reasoning with Language Models:}

The recent successes of large LMs are difficult to understate. However, even the largest-scale and best-performing models continue to struggle with generalizable, interactive reasoning at inference-time. This is particularly true for more complex, longer-horizon, and multi-step tasks \cite{gemini2023, openai2023gpt4, chowdhery2023palm, touvron2023llama, rae2022scaling, brown2020language,radford2019language}. Though supervised finetuning has been demonstrated to improve LM reasoning in certain natural language settings, the problem persists \cite{mishra2022crosstask, zhou2022prompt, lester2021power, chung2022scaling, wang2023far}.

Several methods have looked to abstraction to address these issues. Works like ``Scratchpad'' show that adding hand-engineered annotations to explicitly train language models to reason step-by-step during inference can improve  the quality of generations in multi-step tasks, increasing the cost of inference but requiring no extra demonstration data \cite{nye2021show, li2022language}. 

Other methods including ``Chain-of-Thought'' and  ``Tree-of-Thought''  demonstrate that a well-designed iterative prompting schema can be used \textit{in lieu of finetuning} to systematize reasoning in certain environments \cite{wei2022chain, yao2023tree, besta2023graph}, improving the performance of neural language agents without gradient updates in dodmains like the video game Minecraft \cite{wu2023spring}.

We draw inspiration from both of these lines of work in formulating our method. However, the approach we propose is task-agnostic and requires no hand-crafting, human annotation, or prompt-tuning to improve the quality of generations. Moreover, we extensively test our method on a task that is partially-observable, stochastic, infinite-horizon, and has no compactly summarizable rule-set \cite{kuttler2020nethack, hambro2022insights}, unlike Minecraft \cite{wu2023spring}. 

\paragraph{Natural Language for Supervised Decision-Making:} Many recent works interface natural language and LMs with sequential decision-making. 

\citet{peng2023learning} query humans for descriptions of robot learning tasks and use language models to translate these into task-relevant abstractions for a state-based policy, while \citet{saycan2022arxiv} employ LMs as high-level controllers over low-level skill policies to ground natural language in robot affordances. \citet{jiang2023vima} incorporate text and pixel inputs to train transformer-based vision-language models with supervised learning for table-top manipulation tasks, interleaving tokens across modalities during training. Similarly, \citet{rt22023arxiv} finetune large vision-language models on robot control datasets, expressing robot actions as text. 

Finally, most similar to our work, \citet{li2022language} encode environment goal, history, and visual observations into embeddings, average, and use the results as input to tune a pretrained LM for decision-making in the BabyAI and VirtualHome environments \cite{chevalier2018babyai}. However, both of these settings are low-dimensional compared to the general decision-making settings for which we look to equip language agents in this paper, like NetHack.

\vspace{-1mm}
\paragraph{BabyAI-Text:} BabyAI-Text is a fully natural language-based and multi-task gridworld environment introduced by \citet{carta2023grounding}. It is based on the BabyAI environment developed by \citet{chevalier2018babyai}. 

Each instance of the environment requires an agent to complete a task belonging to one of five classes; to solve each task, agents must interact with procedurally-generated objects and rooms (see Appendix \ref{appendix:babyai}). There has been extensive work studying decision-making in BabyAI \cite{kirk2021survey, reed2022generalist, li2022language}. 
\vspace{-1mm}
\paragraph{NetHack:}
NetHack is an extremely long horizon and open-ended ``dungeon-crawler'' video game. Games typically span from thousands to hundreds of thousands of keypresses, but may go on indefinitely as long as the player avoids death. It is also extremely complex, with players needing to explore and navigate through a series of interlinked ``dungeon'' levels with layout, objects, and monsters randomly regenerated in each instance of the game \cite{kuttler2020nethack, hambro2022insights}. 

As a result, NetHack is very difficult for humans to play, let alone for language  or vision agents. Hand-crafted symbolic policies continue to outperform state-of-the-art neural agents for NetHack by a large margin, including those trained on hundreds of billions of keypresses of expert gameplay \cite{hambro2022insights, hambro2022dungeons, zhong2022improving, klissarov2023motif, piterbarg2023nethack, tuyls2023scaling}. 

We provide an extensive description of NetHack and the NetHack Learning Environment in Appendix \ref{appendix:nethack}.

%% file: method.tex
\begin{figure}[t!]
     \centering
      \includegraphics[width=\linewidth]{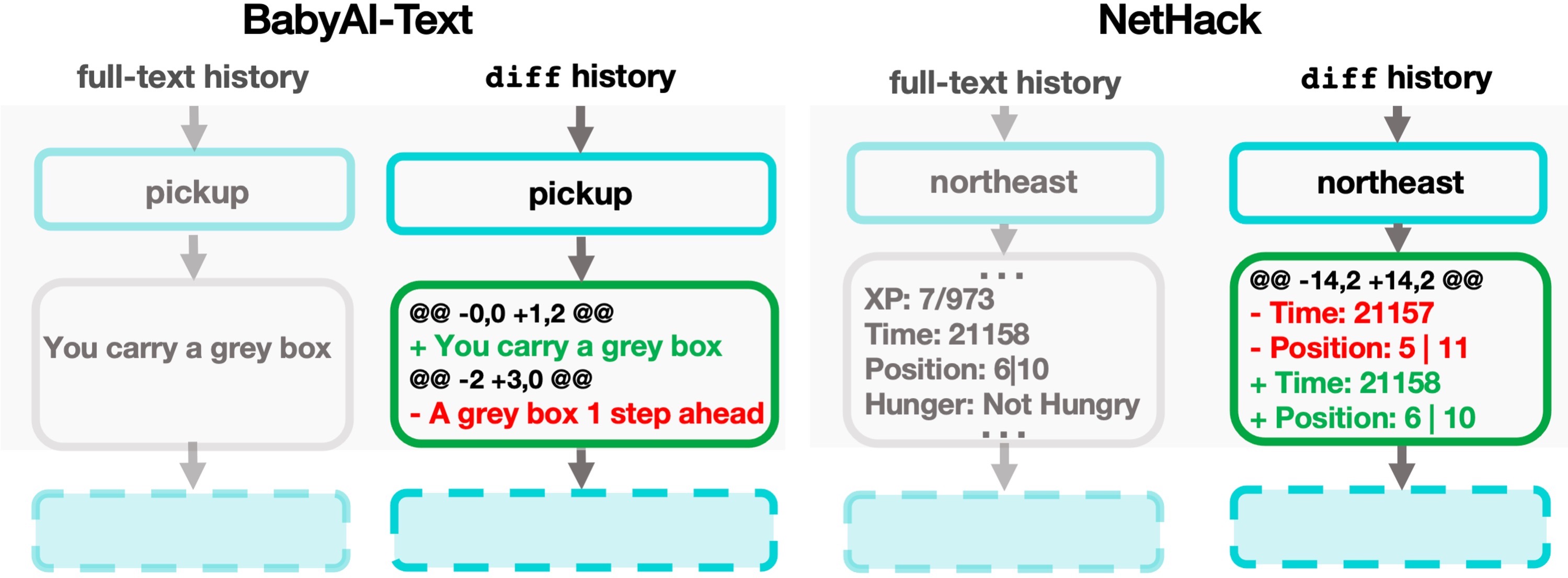}
      \label{fig:diff_visualization}
      \caption{\textbf{A closer look at the text deltas produced by \diff{}}: we provide examples of full-text and \texttt{\small{diff}} observations in our two environments of interest, BabyAI-Text and NetHack. A more detailed example of Unix-style \diff{} outputs can be found in Appendix \ref{appendix:more_on_diff}.}
      \vspace{-2mm}
    \end{figure}
    
\section{\texttt{diff} History}

We look to improve the downstream performance of neural LMs that are trained and/or finetuned for sequential decision-making. Training language models on text-based data may constitute a challenging learning problem, especially in low-sample and high-dimensional settings, where full-text environment observations may be verbose, complex, and information-sparse. To address this, we propose \method{}. In this section, we provide a formalism for \method{}, details on computing it, and a discussion on training LMs with \method{}.

\subsection{Text-Based Decision-Making}
\label{section:diff_hist_tb_dm}
We operate in the text-based imitation learning setting in this work, where we have access to datasets of demonstration data annotated with action labels. Concretely, our training datasets $\mathcal{D}$ consists of sequences of observations and expert actions across contiguous timesteps of interaction, paired with an \textit{instruction} describing the task/goal. A formal expression for $\mathcal{D}$ is provided in Equation \ref{equation:dataset_def}.

\begin{equation}
    \mathcal{D} = \left\{\{g^i, (o^{i}_{1}, a^{i}_{1}, \dots o^{i}_{H}, a^{i}_{H})\}\right\}_{i=1}^{N}
    \label{equation:dataset_def}
\end{equation}

We refer to the interleaved observation and action sequences in $\mathcal{D}$ as \textit{interaction histories}. In our experiments, we train language agents on interaction histories reflecting a fixed, maximal horizon, $H$. 

Because our data are fully text-based, individual instructions, observations, and actions may correspond to multiple tokens. As a result, our finetuning problem is \textit{sequence-to-sequence}: our goal is train neural language agents to effectively model distributions of expert action tokens conditioned on both instruction tokens and interaction history tokens representing some $h: 1 \leq h \leq H$ previous timesteps. 

From a practical standpoint, neural language agents will need to effectively use long histories ($h \gg 1$) to model such distributions in sequential decision-making settings that reveal only partial information at individual timesteps, i.e. partially-observable Markov decision processes \cite{kaelbling1998planning}.

Let $a_{t}^{i}$ be an expert action in $\mathcal{D}$ and let $(o_{t-h+1}^i, a_{t-h+1}^i, \dots, o_{t}^i)$ represent $h$ previous timesteps of interaction reflecting the instruction $g^i$. We can write the tokenized version of this interaction history as shown in Equation \ref{equation:full_text_history_tokens}.

\begin{equation}
    (\boldsymbol{\omega}_{t-h+1}^i, \boldsymbol{\alpha}_{t-h+1}^i, \dots, \boldsymbol{\omega}_{t}^i)
    \label{equation:full_text_history_tokens}
\end{equation}

In the expression above, $\boldsymbol{\omega}_{j}^i$ and $\boldsymbol{\alpha}_{j}^i$ denote the token subsequences corresponding to the observation $o_{j}^{i}$ and the action $a_{j}^{i}$, respectively. Using this notation, a neural LM agent $\pi_{\theta}$ with parameters $\theta$ that models sequences of action tokens can be represented as shown below, with the additional symbol $\boldsymbol{\gamma}^i$ representing the tokenized instruction $g^i$.

\begin{equation}
  \label{equation:dm_agent}
    \begin{aligned}
        \boldsymbol{\alpha}_{t}^i \sim \pi_{\theta}\left(\boldsymbol{\gamma}^i, \boldsymbol{\omega}_{t-h+1}^i, \boldsymbol{\alpha}_{t-h+1}^i, \dots, \boldsymbol{\omega}_{t}^i\right)
    \end{aligned}
\end{equation}

Parameters $\theta$ can be trained or tuned with teacher-forcing  \cite{williams1989learning} by minimizing the cross-entropy loss between predicted next-token logits and one-hot vectors representing ``ground truth'' expert action tokens. We provide a formal and complete definition for this loss function in Appendix \ref{appendix:action_prediction_loss}, Equation \ref{equation:full_text_action_pred}.

 \subsection{Unix  \diff{}s and Text Deltas}
\label{section:diff_hist_defs}

Given a pair of text strings, the Unix operation \texttt{\small{diff}} matches the contents of each string line-by-line by treating each line as a separate substring. It then computes and returns a symbolically annotated summary of any detected differences between the text strings, enumerated line-by-line \cite{kernighan1984unix}. 

The \diff{} operator is implemented in control systems like \texttt{git} and integrated development environments like Visual Studio to help software engineers efficiently track and visualize changes between versions of text files. A detailed example of the outputs and default notation employed by Unix-style \texttt{\small{diff}} is provided in Appendix \ref{appendix:diff_general_example}. 

\subsection{Computing \method{} for Decision-Making}
\label{section:df_hist_computing}

Naively, without \method{}, one might train a model to predict text actions $a_t^i$ for an instruction $g^i$ given an interaction history  $(o_{t-h+1}^i, a_{t-h+1}^i, \dots, o_{t}^i)$ consisting of past full-text observations and actions tokenized as in Equation \ref{equation:full_text_history_tokens}. We refer to such an interaction history as a \textit{full-text history}.

With \method{}, we substitute the full-text observations that follow the first (or ``oldest'') timestep in the history with text deltas $\Delta o_l^i$ by computing Unix \texttt{\small{diff}}'s between consecutive pairs of full-text observations as in Equation \ref{equation:def_diff_history}.

\begin{equation}
    \begin{aligned}
        \Delta o_l^i &\coloneq \textbf{\texttt{\small{diff}}}(o_{l-1}^i, o_l^i), \ \forall l : t-h < l \leq t
    \end{aligned}
    \label{equation:def_diff_history}
\end{equation}

This yields an interaction history that is equivalent to the original (Equation \ref{equation:diff_inter_history}). 

\begin{equation}
    (o_{t-h+1}^i, a_{t-h+1}^i, \Delta o_{t-h+2}^i, a_{t-h+2}^i, \dots, \Delta o_{t}^i)
    \label{equation:diff_inter_history}
\end{equation}

The single, full-text observation $o_{t-h+1}^i$ acts as an ``anchor'' against which subsequent deltas are computed. We refer to interaction histories that are processed using the Unix \texttt{\small{diff}} operation in this manner as \textbf{\texttt{\small{diff}} histories}. To parallel Equation \ref{equation:full_text_history_tokens},  we can write the tokenized version of a \method{} as in Equation \ref{equation:diff_history_tokens}.

\begin{equation}
    (\boldsymbol{\omega}_{t-h+1}^i, \boldsymbol{\alpha}_{t-h+1}^i, \boldsymbol{\delta}_{t-h+2}^i, \boldsymbol{\alpha}_{t-h+2}^i, \dots, \boldsymbol{\delta}_{t}^i)
    \label{equation:diff_history_tokens}
\end{equation}

In the expression above, $\boldsymbol{\delta}_j^i$ denotes the token subsequence corresponding to the delta or \diff{} text observation $\Delta o_j^i$. A \method{} neural LM agent can thus be represented as below. 

\begin{equation}
    \begin{aligned}
        \boldsymbol{\alpha}_{t}^i \sim \pi_{\theta}\left(\boldsymbol{\gamma}^i, \boldsymbol{\omega}_{t-h+1}^i, \boldsymbol{\alpha}_{t-h+1}^i, \boldsymbol{\delta}_{t-h+2}^i, \dots, \boldsymbol{\delta}_{t}^i\right)
    \end{aligned}
\end{equation}

The loss function that we employ for training or finetuning $\theta$ on \diff{} histories is defined in Appendix \ref{appendix:action_prediction_loss}, Equation \ref{equation:diff_action_pred}. A diagrammatic visualization of \method{} is provided in Figure \ref{fig:intro_figure} (right) as well as in Figure \ref{fig:diff_visualization}, where we show examples of the annotated summaries of differences between consecutive full-text observations output by \diff{}. 

\subsection{Why would \method{} help?}
\label{section:diff_hist_whys}

We conjecture two possible effects of \method{}: (1) the addition of task-agnostic and relational abstraction to sequence modeling for decision-making; (2) token-level compression of long interaction histories in high dimensional environments.

\paragraph{Task-Agnostic and Relational Abstraction:} The Unix \texttt{\small{diff}} operation computes pairwise text deltas between arbitrary strings. This makes it task-agnostic. Moreover, text deltas computed with \texttt{\small{diff}} are structured like code, providing a summary of text differences using common-place symbols, notations, and abstractions via a simple ``arithmetic-style'' logic, as demonstrated in Figure \ref{fig:diff_visualization}, Appendix \ref{appendix:diff_general_example}, and Appendix \ref{appendix:interaction_history_examples}. Thus, finetuning LMs for sequential decision-making with \texttt{\small{diff}} observations does not require re-tokenization or the introduction of large sets of new tokens. This property makes it attractive for settings with little labeled data. Finally, as shown in Figure \ref{fig:diff_visualization}, the line-by-line matching performed by \diff{} can localize salient changes to environments that occur between consecutive timesteps of interaction, tracking changes in objects' states and relations, high level features of the scene, and more.
As a result, \method{} may provide a denser and richer learning signal during LM finetuning than the one provided by ``raw'' full-text interaction histories, improving downstream agent generalization after finetuning on limited data.

\paragraph{Token-Level Compression:} Language model context length defines a maximal ``computational budget'' for token generation. Compressing relevant historical information as densely as possible on the token-level across context-length thus offers a direct means for extending and improving neural agent memory. This may result in improved modeling of expert action distributions during instruction finetuning, yielding LMs that produce higher quality generations at test-time. We explicitly probe this property in our experiments (see Figure \ref{fig:token_stats} and Section \ref{subsection:diff_compresion}).

%% file: results.tex
\section{Experiments}

To understand the benefits of \method{}, we experiment on two different procedurally-generated environments: BabyAI-Text \cite{chevalier2018babyai, carta2023grounding} and NetHack \cite{kuttler2020nethack}. Concretely, we seek to answer the following questions:

\begin{enumerate}[leftmargin=*]
    \item How does \method{} impact the quality of generations after finetuning LMs on limited labeled data? How does it impact the robustness of neural language agents?
    
    \item Does large-scale task-agnostic pretraining affect the performance of \method{} neural language agents?
    \item Does the downstream performance of \method{} agents scale with data samples?
    \item How do \method{} agents compare against state-of-the-art neural agents?
\end{enumerate}

We probe (1) in both BabyAI-Text and NetHack. In the remainder of our study, we concentrate our investigations on the high-dimensional environment NetHack, where low-resource neural agent training and finetuning have been demonstrated to be exceptionally difficult in prior work \cite{hambro2022insights, hambro2022dungeons, tuyls2023scaling, piterbarg2023nethack}.
Summaries of our experimental results are provided in Figures \ref{fig:babyai_learning_curves}, \ref{fig:nethack_learning_curves}, and Table \ref{tab:agg_nethack_results}.

\begin{figure}[b!]
     \centering
        \vspace{-2mm}
      \includegraphics[width=0.7\linewidth]{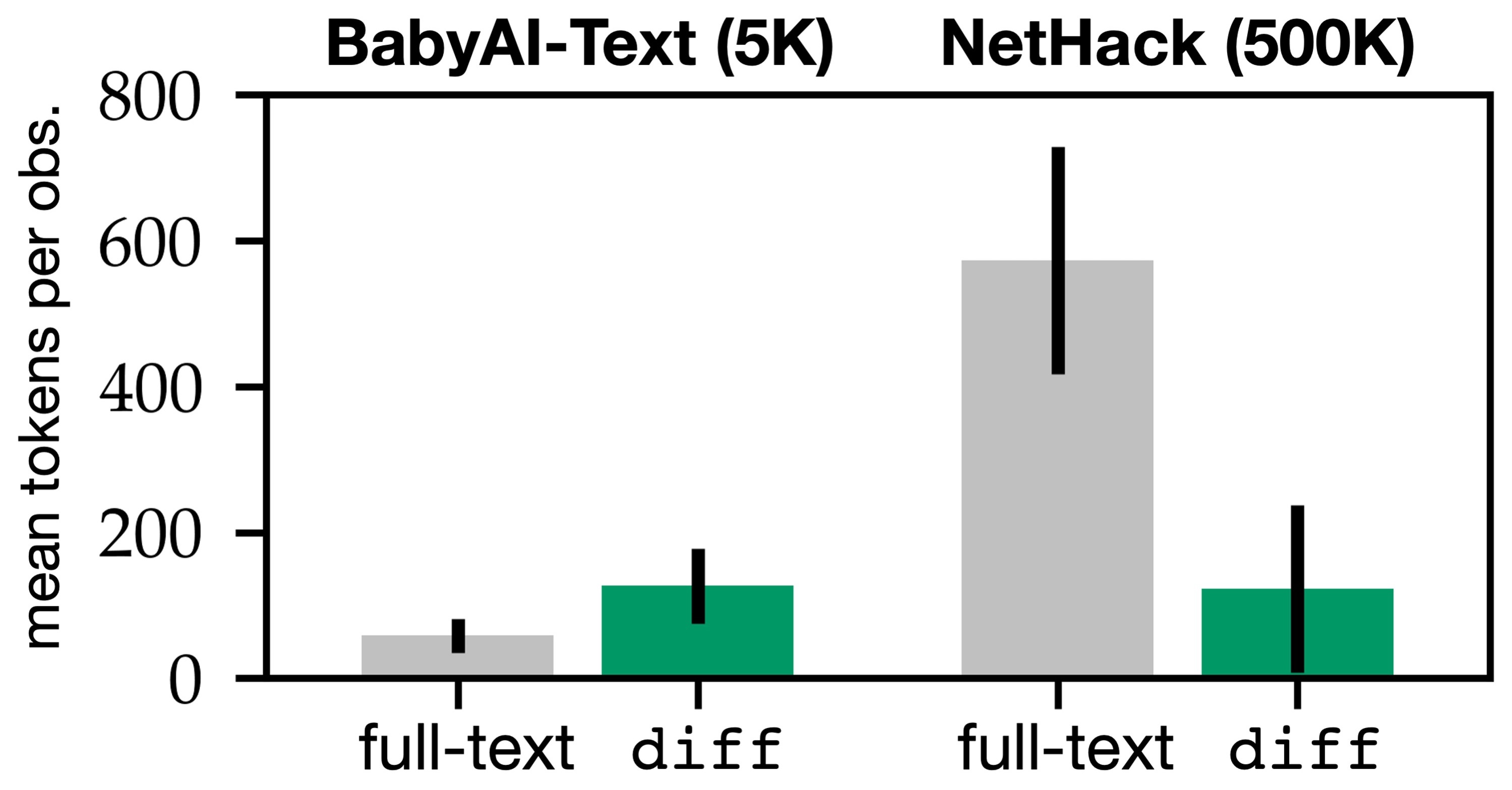}
      \caption{\textbf{Mean token counts per observation, before and after \texttt{diff}}: We compute token statistics of full-text and \texttt{\small{diff}} observations across natural language interaction datasets generated for a low-dimensional environment (BabyAI-Text) and a high-dimensional environment (NetHack) via the tokenizer we use for all LM experiments \cite{radford2019language}. Error bars indicate one standard deviation across datasets.}
      \label{fig:token_stats}
    \end{figure}

\subsection{Generating Natural Language Interaction Data}
\label{section:exp_data}

\subsubsection{BabyAI-Text}
We leverage the procedural BabyAI bot \cite{chevalier2018babyai} to collect interaction demonstrations in the multi-task BabyAI-Text environment. The configuration of this environment is as in \citet{carta2023grounding}, so that data reflect five classes of tasks. A description of each of these task classes is provided in Appendix \ref{appendix:babyai}.

We first run the bot across distinct random seeds of the environment to collect 5,000 full-length goal-conditioned solutions. Then, we sub-sample these across ``chunks'' $(H = 4)$ to yield two low-sample datasets, consisting of 1,000 and 5,000 goal-conditioned interaction demonstrations respectively. Further data statistics are reported in Appendix \ref{appendix:babyai_dataset}.

\begin{figure*}[ht]
    \centering
    \includegraphics[width=0.93\textwidth]{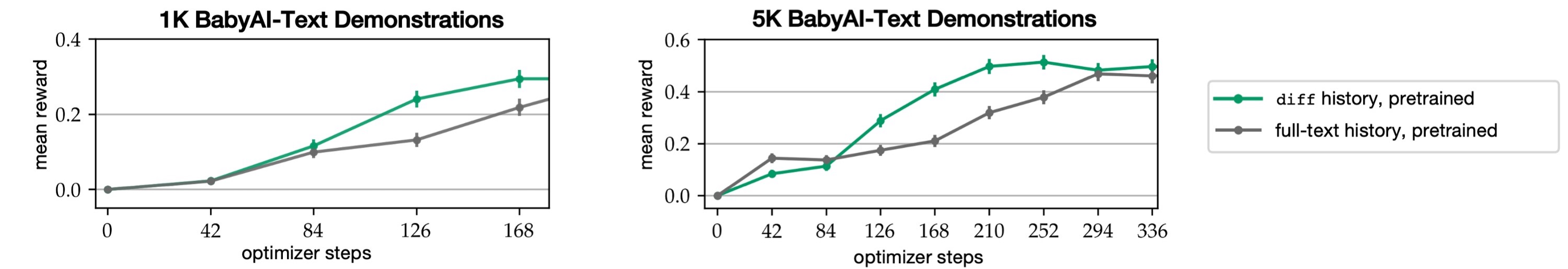}
    \vspace{-1mm}
    \caption{\textbf{Low-resource BabyAI-Text}: We compare the performance of  language agents tuned with \method{} against full-text history ablations in two low-sample finetuning settings, with 1,000 demonstrations (left) and 5,000 demonstrations (right) respectively. Though both agents eventually achieve equivalent performance over the course of finetuning in both experiments, the \method{} language agent requires \textbf{$\geq$25\% fewer FLOPs} to attain this peak. Error bars indicate one standard error in mean reward on a batch of withheld task instances.}
    \label{fig:babyai_learning_curves}
\end{figure*}

\subsubsection{NetHack}
\label{section:langhack}

While the video game NetHack is vision-based by default, the recent NetHack Language Wrapper \cite{goodger2023nethack} makes it possible to play the game fully through natural language. Unfortunately, as of the writing of this paper, existing datasets for NetHack are incompatible with this wrapper. These datasets store games in the impoverished \texttt{\small{tty*}} format, which makes it difficult to procedurally generate natural language descriptions of visual observations \cite{kuttler2020nethack, hambro2022insights, piterbarg2023nethack, tuyls2023scaling}. 

Thus, to support our experiments with \method{}, we generate a new NetHack dataset consisting of full game states by running the leading symbolic bot \texttt{\small{AutoAscend}} in the environment \cite{hambro2022insights}. Note that all current state-of-the-art neural agents for NetHack are trained or pretrained on large datasets of \texttt{\small{AutoAscend}} games via supervised learning from pixels \cite{hambro2022dungeons, piterbarg2023nethack, tuyls2023scaling}. We call our dataset \dataset{}. 

Our procedure for generating \dataset{} mirrors the procedure that we use in BabyAI-Text. We first run the \texttt{\small{AutoAscend}} bot across 10,000 NetHack seeds, recording full-length games. Then, we sub-sample these games uniformly at random into a dataset appropriate for LM finetuning consisting of 500,000 ``chunks'' of interaction demonstrations $(H = 64)$. Note that we scale up the length of NetHack interaction histories compared to the length of the interaction histories that we employ for BabyAI-Text data experiments $(H=4)$ due to the difference in difficulty between these two environments. A detailed report of \dataset{} statistics is provided in Appendix \ref{appendix:langhack}

\subsubsection{Properties of generated \method{} data}
\label{subsection:diff_compresion}

We probe the effect of \method{} on the token statistics associated with environment observations in Figure \ref{fig:token_stats}. In NetHack, we find that \texttt{\small{diff}} acts as a  ``soft compression'' mechanism on natural language observations, yielding a 460\% reduction in average token count per timestep. This finding supports our conjecture in Section \ref{section:diff_hist_whys}. Nevertheless, we find that \diff{} has \textit{the opposite effect} in BabyAI-Text, where it produces a 210\% average-case increase in per observation tokens. We attribute this property to the difference in state dimensionality between the two environments. Indeed, inspection of the generated data reveals that the natural language observations in BabyAI bot demonstrations change entirely between most consecutive timesteps of interaction. However, the relational property of \diff{} described in Section \ref{section:diff_hist_whys} holds in both environments: \diff{} observations consistently identify and localize per timestep changes in the descriptions of objects and scene properties. Examples of BabyAI-Text and NetHack \diff{} interaction histories are provided in Appendix \ref{appendix:interaction_history_examples}. 

\subsection{\texttt{diff} vs. Full-Text History Language Agents}
\label{section:diff_vs_fulltext}
We begin our investigations by comparing the performance of neural language agents trained with \method{} against full-text history ablations in BabyAI-Text and NetHack as a function of optimizer steps.

In all experiments, we tune the smallest parameter-count checkpoint of the decoder-only language model GPT-2 \cite{radford2019language}, employing identical hyperparameters and hardware across individual environments. Finetuning is conducted in mixed-precision with Microsoft DeepSpeed \cite{wang2023zero++} and HuggingFace Accelerate \cite{accelerate} on NVIDIA A100 GPU nodes. We preserve the default 1024-token context length of the model in BabyAI-Text, but extend model context lengths by a factor of four up to 4096 tokens for finetuning on \dataset{} by introducing new positional encodings. All samples are padded or truncated to these context lengths, though we note that we find truncation to be unnecessary on our low-resource BabyAI-Text interaction datasets. A complete description of our training procedures is provided in Appendix \ref{appendix:training_details}.

To evaluate agents, we run a batch of games on withheld, procedurally-generated instances of both environments. At each timestep of interaction, we provide recent interaction history in-context to LMs (as visualized in Figure \ref{fig:intro_figure}) and use greedy decoding \cite{lowerre1976harpy} to generate natural language actions. We use no token sampling restrictions or other heuristics during conditional generation. Following previous work, we employ the total reward and/or  game score achieved by agents across withheld task instances as a metric for language agent performance in our evaluations. More details on evaluation are in Appendix \ref{appendix:eval_details}.

We find the introduction of \method{} to result in significant improvements in neural language agent performance in all of our instruction finetuning experiments. Learning curves reflecting the results of \texttt{\small{diff}} ablations on neural agent performance  are displayed in Figure \ref{fig:babyai_learning_curves} and Figure \ref{fig:nethack_learning_curves}, respectively. We employ the loss functions defined in Appendix \ref{appendix:action_prediction_loss} and Appendix \ref{appendix:action_world_model_prediction_loss} to tune models.

\begin{figure*}[ht!]
    \centering
    \includegraphics[width=0.92\textwidth]{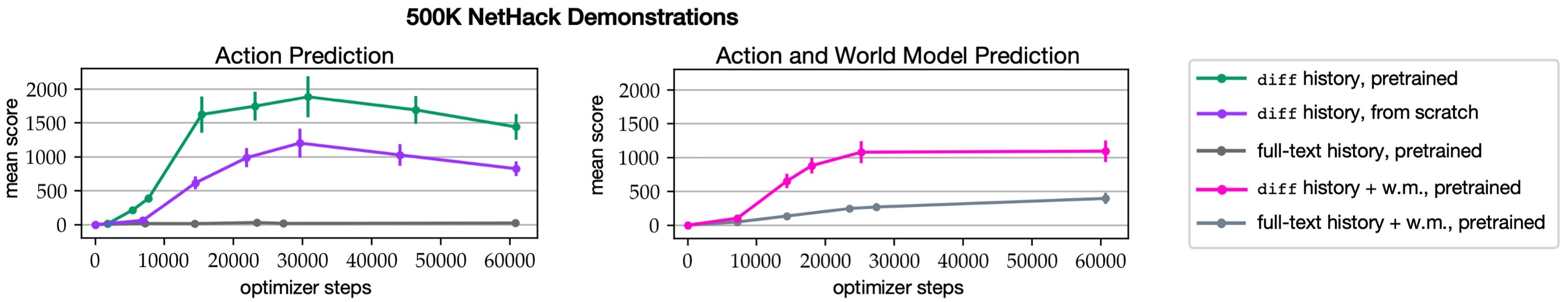}
    \vspace{-1mm}
    \caption{
    \textbf{Low-resource NetHack}: We compare the performance of language agents with \method{} against ablations, testing efficiency and generalization when LMs are tuned on demonstrations of interaction history for natural language action prediction (left) and joint action-world model prediction (right). Error bars indicated one standard error in mean game score achieved by agents across a batch of withheld NetHack instances.
    }
    \label{fig:nethack_learning_curves}
\end{figure*}

\subsubsection{Action Prediction}
\label{section:results_action_pred}
In BabyAI-Text, we find that ablating \texttt{\small{diff}} observations when finetuning language agents for action prediction  does not impact the eventual performance achieved by models. However, \method{} agents reach this performance level with more than 25\% fewer optimizer steps or FLOPs in both the ultra low-sample and low-sample labeled data settings that we test (Figure \ref{fig:babyai_learning_curves}). We note that this improvement in efficiency occurs despite the average-case token count increase effect identified in Section \ref{subsection:diff_compresion} and Figure \ref{fig:token_stats}.

In NetHack, the effect of ablating \texttt{\small{diff}} observations from low-sample finetuning on action prediction is more dramatic, resulting in a 98\% drop in the final max-performance achieved by neural language agents on new game seeds (see Figure \ref{fig:nethack_learning_curves} (left) and Table \ref{tab:agg_nethack_results}).

\subsubsection{Action and World Model Prediction}
\label{section:res_text_world_model}
In an effort to probe the performance gap  between \texttt{\small{diff}} and full-text history agents tuned on \dataset{}, we experiment with removing all masking from observation tokens in the demonstration data used for finetuning. This results in language agents tuned not only on action prediction, but also on an \textit{auxiliary natural language world modeling objective}. The conjecture behind this experiment is that in the information-sparse, full-text NetHack setting, introducing an auxiliary objective may improve the efficiency of representation learning. Our findings are displayed in Figure \ref{fig:nethack_learning_curves} (right).

Indeed, we find that the addition of the causal world model objective narrows the gap between LMs tuned with \diff{} and full-text history, reducing the 98\% difference in agent mean NetHack score on test games to a 69\%  difference  (see Table \ref{tab:agg_nethack_results}). The auxiliary objective produces a large improvement in the quality of full-text history agent generations (10-fold improvement in mean NetHack score), but decreases the quality of \method{} generations. Finetuning LMs with \method{} data on action prediction alone yields stronger representations for decision-making in NetHack.

\subsection{Large-Scale Pretraining}
\label{section:pretraining}
Next, we test the effect of ablating large-scale pretraining from \method{} language agents in NetHack. Learning curves reflecting this ablation are also displayed in Figure \ref{fig:nethack_learning_curves} (left).  We find that pretraining \textit{does} have a significant positive effect on the performance achieved by \method{} NetHack agents -- re-initializing model weights to train on \dataset{} from scratch results in a 36\% reduction in mean NetHack game score (see Table \ref{tab:agg_nethack_results}). Interestingly, we also find that the performance gap between pretrained and from-scratch agents remains fairly constant with optimizer steps over the course of training/tuning (see Appendix \ref{appendix:training_details}).

\subsection{Doubling NetHack Interaction Demonstrations}
\label{section:results_scaling}
We confirm that finetuning LMs on labeled interaction demonstrations with \method{} scales with data by running an experiment in which we double the number of samples in \dataset{}. We refer to this upscaled dataset as \texttt{\small{LangHack-2x}}. All other properties of \texttt{\small{LangHack-2x}} are precisely the same as those of \dataset{} described in Section \ref{section:langhack}, e.g. interaction demonstration ``chunks'' ($H=64$) are sampled from the same set of 10,000 full \texttt{\small{AutoAscend}} games. The high-level statistics of \texttt{\small{LangHack-2x}} match those of \texttt{\small{LangHack}} (see Table \ref{tab:agg_data_stats}).

The results from this experiment are summarized in Table \ref{tab:agg_nethack_results} and in Figure \ref{fig:nethack_upscaled}. Agents trained with \method{} for action prediction on \texttt{\small{LangHack-2x}}  achieve 47\% higher mean NetHack score than those trained on \dataset{}.

\begin{table*}[t!]
\caption{\textbf{Summary of experimental results on NetHack}: We compare the best in-game performance achieved by neural vision and language agents trained on \dataset{} and \texttt{LangHack-2x} against previous state-of-the-art neural agents for NetHack. These agents are trained on orders of magnitude more game data from the leading NetHack symbolic bot \texttt{\small{AutoAscend}} \cite{hambro2022insights}. We annotate agent scores statistics from \citet{piterbarg2023nethack} with $(\dagger)$ and those from \citet{tuyls2023scaling} with $(\ddagger)$. Details on evaluation procedures are provided in Appendix \ref{appendix:eval_details}. Models with architecture denoted as ``LSTM (CDGPT)'' and ``XL-LSTM (CDGPT5)'' reflect the vision-language model architecture for NetHack introduced as ``Chaotic Dwarven GPT-5'' (CDGPT5) in the NeurIPS 2021 NetHack Challenge Competition, which outperformed all other purely data-driven polices \cite{hambro2022insights}.}

\label{tab:agg_nethack_results}
\begin{center}
\begin{small}
\begin{sc}

\begin{tabular}{ ccccccccc }

\toprule
&& & & & &  \multicolumn{3}{c}{\textbf{NetHack Performance}}\\

\cmidrule(r){7-9} \shortstack{Agent \\ Type} & \shortstack{Model \\ Architecture} & \shortstack{\# of \\ Expert \\ Actions} & \shortstack{Data\\ Format} & \shortstack{World\\ Model} & \shortstack{Weight \\ Init} & \shortstack{Best \\ History \\Horizon} & \shortstack{Mean\\ Score} &  \shortstack{Median\\ Score} \belowspace\\
\hline \hline
\abovespace
Vision & \scriptsize{LSTM (CDGPT5)} & 32M & \scriptsize{Pixel-Vector}& - & \scriptsize{Scratch}  & inf & 192 $\pm$ 7 & 111 \\
Vision &  Trnsfrmr & 32M &  \scriptsize{Pixel-Vector} & - & \scriptsize{Scratch} & 2 & 213 $\pm$ 6 & 114 \belowspace \\
\abovespace
Language & GPT-2-127M & 32M & \scriptsize{Full-Text} & \xmarkb & \scriptsize{Pretrain} & 8 & 31 $\pm$ 5 & 20\\ 
\belowspace  Language& GPT-2-127M & 32M & \scriptsize{Full-Text} & \cmarkb & \scriptsize{Pretrain} & 8 & 344 $\pm$ 52 & 171
\\
\abovespace 
Language & GPT-2-127M & 32M & \scriptsize{\texttt{diff} History} & \xmarkb &\scriptsize{Pretrain} & 15& \textcolor{diffgreen}{\textbf{1885 $\pm$ 213}} & 1081 \\
Language & GPT-2-127M & 32M & \scriptsize{\texttt{diff} History} & \xmarkb  & \scriptsize{Scratch} & 15 & 1203 $\pm$ 214 & 645 \\
Language & GPT-2-127M & 32M & \scriptsize{\texttt{diff} History} & \cmarkb & \scriptsize{Pretrain} & 15 & 1094 $\pm$ 160 & 453 \belowspace \\
Language & GPT-2-127M & 64M & \scriptsize{\texttt{diff} History} & \xmarkb &\scriptsize{Pretrain} & 15&
\textcolor{diffgreen_scaled}{\textbf{2766 $\pm$ 481}} & 1802  \belowspace \\
 \hline \hline \abovespace Vision & \scriptsize{LSTM (CDGPT5)} & 3.2B & \scriptsize{Pixel-Vector} & -& \scriptsize{Scratch}  & inf & $\text{658}^{(\dagger)}$ & $\text{403}^{(\dagger)}$\\
Vision & \scriptsize{Trnsfrmr-LSTM} & 3.2B & \scriptsize{Pixel-Vector} & - & \scriptsize{Scratch}  & inf & $\textbf{1318}^{(\dagger)}$& $917^{(\dagger)}$\\
Vision & \scriptsize{XL-LSTM (CDGPT5)}& 115B & \scriptsize{Pixel-Vector} & - & \scriptsize{Scratch} & inf & $\textbf{2740}^{(\ddagger)}$ & - \belowspace \\
\bottomrule
\end{tabular}
\end{sc}
\end{small}
\end{center}
 
\end{table*}

\subsection{\method{} Language Agents vs. NetHack SOTA }
\label{section:exp_abl_res}
Finally, we compare the generalization of NetHack language agents against that of: (1) state-of-the-art vision agents trained for action prediction on \dataset{}; and (2) overall state-of-the-art neural agents for NetHack. Our results are summarized in Figure \ref{fig:intro_figure} (left), Table \ref{tab:agg_nethack_results}, and Figure \ref{fig:nethack_vision_learning_curves}.

To support (1), we render demonstrator observations in \dataset{} in pixels via the procedures used in previous works \cite{hambro2022insights}. We train two types of vision agents for NetHack: the popular LSTM-based \texttt{\small{CDGPT5}} architecture developed during the NeurIPS 2021 NetHack Challenge Competition as well as a Transformer-based variant of \texttt{\small{CDGPT5}}, where the core LSTM module of the network is swapped for a Transformer as in \citet{piterbarg2023nethack} (see Appendix \ref{appendix:nethack_vision_model}). All of our language agents outperform state-of-the-art vision-based counterparts \cite{hambro2022dungeons, piterbarg2023nethack,tuyls2023scaling}, with the exception of the full-text history action prediction agent.

As for (2), we find that \method{} agents tuned on \dataset{} outperform neural vision-language agents trained on $100\times$ more data from the same demonstration source, the symbolic bot \texttt{\small{AutoAscend}}. Furthermore, our \method{} language agent tuned for action prediction on \texttt{\small{LangHack-2x}} matches the performance of the overall current state-of-the-art neural agent for NetHack from \citet{tuyls2023scaling}, which is trained on $1800\times$ as many labeled \texttt{\small{AutoAscend}} samples. The NetHack score distribution of our best agent has a long tail; indeed, the LM achieves a score exceeding $15000$ (see Appendix \ref{appendix:scaling}, Figure \ref{fig:nethack_upscaled_distribution}) and descends beyond dungeon level six in several test games.

%% file: discussion.tex
\section{Discussion, Limitations, and Conclusion}
We propose a simple technique that uses the Unix \texttt{\small{diff}} operator to dramatically improve the quality of LM generations for sequential decision-making. \method{} provides a procedural, deterministic, task-agnostic, and human-interpretable way to reformulate the interaction histories used to prompt models, removing redundancies between consecutive observation pairs in favor of a summary of salient differences in objects, scenes, and other abstract entities. It can also extend the modeling horizon of language agents at fixed context lengths. Through our experiments with \method{}, we demonstrate by example that introducing abstraction into the prompts used to finetune language models into agents can make a powerful impact on downstream performance. 

In BabyAI-Text, our experiments show that \method{} yields a 25\% improvement in the computational efficiency of low-data instruction finetuning over full-text interaction history ablations (Figure \ref{fig:babyai_learning_curves}).

In the infinite-horizon video game NetHack, \method{} makes it possible to tune tiny LMs (127 million parameters) into exceptionally strong agents using relatively little labeled data, but many tuning epochs (Appendix \ref{appendix:training_details}). Under our paradigm, neural language agents achieve state-of-the-art performance for neural agents on NetHack with 1800$\times$ fewer labels than prior work (Figure \ref{fig:intro_figure}). The results of our experiments are reported in Figure \ref{fig:intro_figure}, Figure \ref{fig:nethack_learning_curves}, and Table \ref{tab:agg_nethack_results}. They reveal three key effects:
\vspace{-1mm}
\begin{enumerate}
    \item Language agents with \method{} outperform vision-language baselines trained on equivalent data (89\% higher mean NetHack score).

    \item Ablating \texttt{\small diff} observations from the language model contexts used for finetuning and prompting destroys language agent performance (98\% lower mean NetHack score).
    \item The performance of \method{} language agents scales with data. Tuning on tokens corresponding to twice as many expert actions results in a 46\% improvement in mean score, suggesting that the performance of NetHack \method{} agents remains data-limited.
\end{enumerate}

\vspace{-3mm}
There are several limitations to our work, which if addressed, can bring us closer to general-purpose neural language agents. 

First, we assume that observations and actions in decision-making can be expressed completely in textual form. Scaling to arbitrary problems will require work in procedurally converting observations to text or efficiently combining LMs with other modalities. Recent methods exploring this line of research in robot learning settings suggest that it holds promise \cite{peng2023learning, rt22023arxiv, liu2024ok}. 

Second, the abstractions introduced by \method{} exploit structure that is present apriori in textual observations. This makes \method{} particularly compelling for textual tasks with inherent observational structure, such as web navigation, code editing, and the procedural game settings that we test here. However, we do not expect \method{} to provide improvement to LM agents in settings with entirely unstructured and ``narrative-like'' text observations. Nevertheless, we conjecture that it may be possible to make such environments more amenable to \method{} through the addition of a procedural pre-processing step; for example, by using a parser to convert observations into an object- or agent-centric ``stateful'' form. 

Finally, our best language agents do not utilize explicit planning, a la System 2 inference \cite{kahneman2011thinking}. Combining the ability of \method{} to understand historical observations with explicit modeling of future ones will be an exciting direction for future work. 

\section*{Acknowledgements}
This work is supported by ONR award numbers N00014-21-1-2404 and N00014-21-1-2758. UP is supported by the NSF GRFP Fellowship. LP is supported by the Packard Fellowship. We are especially grateful to Shenglong Wang and NYU High Performance Computing for their support. 

\section*{Impact Statement}
This paper presents work whose goal is to advance the field of Machine Learning. There are many potential societal consequences of our work, none which we feel must be specifically highlighted here.

%% file: appendix.tex
\appendix
\onecolumn

\section{More on \texttt{diff} History}
\label{appendix:more_on_diff}

\subsection{A Simple Example}
\label{appendix:diff_general_example}
Let  $(\texttt{\small{file\_i.txt}}, \texttt{\small{file\_j.txt}})$ be a pair of text files. The Unix operation \texttt{\small{diff}} computes the \textit{delta} between these files. For example, suppose that \texttt{\small{file\_i.txt}} contains the text shown below as its contents.
\begin{equation*}
    \begin{aligned}
         &\texttt{\small{Orange}}\\
        &\texttt{\small{Banana}}\\
        &\texttt{\small{Mango}}
    \end{aligned}
\end{equation*}

Now, suppose that \texttt{\small{file\_j.txt}} has the contents below.
\begin{equation*}
    \begin{aligned}
         &\texttt{\small{Orange}}\\
        &\texttt{\small{Apple}}\\
        &\texttt{\small{Mango}}
    \end{aligned}
\end{equation*}

Then, executing the command ``\texttt{\small{diff}} \texttt{\small{file\_i.txt}} \texttt{\small{file\_j.txt}}'' in Unix will return an output that looks similar to the following.
 \begin{equation*}
     \begin{aligned}
         &\texttt{\small{--- file\_i}}\\
          &\texttt{\small{+++ file\_j}}\\
          &\texttt{\small{@@ \ -1,2 \ +1,2 @@}} \\ 
           &\texttt{\small{-Banana}}\\
            &\texttt{\small{+Apple}}
     \end{aligned}
 \end{equation*}

 As demonstrated in this example, the text deltas computed by the Unix \texttt{\small{diff}} operation omit information\footnote{In some implementations of \texttt{\small{diff}}, a few lines that remain unchanged between files but that are \textit{adjacent} to additions or deletions may also returned by \texttt{\small{diff}} as ``context.'' However, in many implementations of  Unix-style \texttt{\small{diff}} such as the one provided by the Python library \texttt{\small{difflib}}, such ``context'' can be turned off.} about lines that remain unchanged between files, i.e. the lines in \texttt{\small{file\_i}} with an exact match in \texttt{\small{file\_j}}, summarizing additions and deletions only. The \texttt{\small{diff}} operation is not commutative.

\subsection{Action Prediction Losses}
\label{appendix:action_prediction_loss}

As in Section \ref{section:diff_hist_defs}, let $\{g, (o_{1}, a_{1}, \dots, o_{H}, a_{H})\}$ be a full-length interaction history sample reflecting a task instruction $g$ from a demonstration dataset $\mathcal{D}$ for sequential decision-making. Following the notation introduced in the main paper, we write the tokenized version of the instruction as the sequence $\boldsymbol{\gamma}$. Similarly, we write the tokenized \textit{full-text} version of the interaction history in terms of token subsequences as
\begin{equation*}
    (\boldsymbol{\omega}_{1}, \boldsymbol{\alpha}_{1}, \dots, \boldsymbol{\omega}_{H}, \boldsymbol{\alpha}_{H}),
\end{equation*}

and the \textit{\method{}} version as,
\begin{equation*}
    (\boldsymbol{\omega}_{1}, \boldsymbol{\alpha}_{1}, \boldsymbol{\delta}_{2}, \boldsymbol{\alpha}_{2}, \dots, \boldsymbol{\delta}_{H}, \boldsymbol{\alpha}_{H}).
\end{equation*} 

In the expressions above, the symbols $\boldsymbol{\omega}_i$ denote token subsequences corresponding to full-text observations, $\boldsymbol{\alpha}_i$ denote token subsequences corresponding to actions, and $\boldsymbol{\delta}_i$ denote token subsequences corresponding to \diff{} observations.

Next, we borrow some notation from computability theory \cite{sipser1996introduction}. Let $\boldsymbol{\Sigma}$ denote the set of all defined tokens, or the \textit{language} of our modeling problem. In the settings that we consider, the tokens used to represent observation and action data for sequential decision-making are intentionally chosen to belong to the same language, i.e. all $\boldsymbol{\omega}$, $\boldsymbol{\alpha}$, and $\boldsymbol{\delta}$ are sequences of elements in $\boldsymbol{\Sigma}$, and individual tokens may very well be reused between these different parts of interaction history. 

We can further write out token subsequences in terms of their constituent tokens as $\boldsymbol{\omega}_{i} = (\omega_{i}^{1}, \dots)$, $\boldsymbol{\alpha}_{i} = (\alpha_{i}^{1}, \dots)$, $\boldsymbol{\delta}_{i} = (\delta_{i}^{1}, \dots)$, where all $\omega_{i}^{j}, \alpha_{i}^{j}, \delta_{i}^{j} \in \boldsymbol{\Sigma}$.

The loss functions that we minimize when tuning LMs with parameters $\theta$ on \textbf{causal action prediction} from interaction history sequences can thus be written as,

\begin{align}
    \mathcal{L}_{\text{action-full-text}} &= - \sum_{i} \sum_{j}\log p_{\theta}\left(\alpha_i^{j} \ | \ \boldsymbol{\gamma}, (\boldsymbol{\omega}_1, \boldsymbol{\alpha}_1, \dots, \boldsymbol{\omega}_{i}), (\alpha_{i}^{1}, \dots \alpha_{i}^{j-1})\right)
    \label{equation:full_text_action_pred}
\end{align}

\begin{align}
    \mathcal{L}_{\text{action-\texttt{diff}}} &= - \sum_{i} \sum_{j}\log p_{\theta}\left(\alpha_i^{j} \ | \ \boldsymbol{\gamma}, (\boldsymbol{\omega}_1, \boldsymbol{\alpha}_1,  \boldsymbol{\delta}_2, \dots, \boldsymbol{\omega}_{i}), (\alpha_{i}^{1}, \dots \alpha_{i}^{j-1})\right).
    \label{equation:diff_action_pred}
\end{align}

In other words, we look to maximize the likelihood of individual expert action tokens $\alpha_i^{j}$ conditioned on: (1) the task instruction; (2) all interaction history up to the timestep $j$; and (3) the $(j-1)$-length prefix of $\boldsymbol{\alpha}_i$. Note that these objectives reflect the use of teacher-forcing \cite{williams1989learning}.

\subsection{World Model Prediction Losses}
\label{appendix:action_world_model_prediction_loss}
The definitions of the \textbf{auxiliary causal world model prediction} losses for full-text and \method{} neural language agents (as explored in Section \ref{section:res_text_world_model} of our experiments) mirror the losses above. 

\begin{align}
    \mathcal{L}_{\text{world-full-text}} &= - \sum_{i \ :\  i\ >\ 1} \sum_{j}\log p_{\theta}\left(\omega_i^{j} \ | \ \boldsymbol{\gamma}, (\boldsymbol{\omega}_1, \boldsymbol{\alpha}_1, \dots, \boldsymbol{\omega}_{i-1}), (\omega_{i}^{1}, \dots \omega_{i}^{j-1})\right)
\end{align}

\begin{align}
\mathcal{L}_{\text{world-\texttt{diff}}} &=  - \sum_{i \ :\  i\ >\ 1} \sum_{j}\log p_{\theta}\left(\delta_i^{j} \ | \ \boldsymbol{\gamma}, (\boldsymbol{\omega}_1, \boldsymbol{\alpha}_1,  \dots, \boldsymbol{\delta}_{i-1},\boldsymbol{\alpha}_{i-1}), (\delta_{i}^{1}, \dots \delta_{i}^{j-1})\right).
\end{align}

Minimizing these objectives is equivalent to maximizing the likelihood of observation tokens in generated full-text and \diff{} history demonstration data conditioned on: (1) the task instruction; (2) all previous interaction history; and (3) the $(j-1)$-length prefix of $\boldsymbol{\omega}_i$ or $\boldsymbol{\delta}_i$. As above, our world model prediction losses use teacher-forcing.

\newpage

\section{Formatting Interaction Histories}
\label{section:diff_hist_format}
\subsection{Beginning and End-of-Subsequence Tokens}

As noted in Section \ref{section:diff_hist_tb_dm}, the lengths of the token subsequences corresponding to individual observations and actions in interleaved interaction histories might vary from timestep to timestep. Thus, we need a mechanism to distinguish between observation and action token subsequences. We draw inspiration from the formatting methods employed by instruction tuning approaches to tune LMs for chat \cite{wang2023far} to resolve this problem, introducing two special tokens to LM tokenizers: $\texttt{\small{<|action|>}}$ and $\texttt{\small{<|observation|>}}$. 

These special tokens are added to the start and end of each action token subsequence, respectively. This choice allows us to condition LM policies on interaction histories with variable-length horizons, maneuvering around the constraint of fixed LM context-lengths to ``look back'' as many timesteps $t$ into the past as we can fit in-context up to a horizon $h$, where $t: 1\leq t \leq h \leq H$ and $H$ represents the maximal horizon of interaction histories represented in the demonstration dataset $\mathcal{D}$ (see Section \ref{section:df_hist_computing}).

We add these tokens to tokenizers at the start of training or tuning, resizing pre-configured language model token embedding dimensions accordingly.

\subsection{Considerations During Generation}
\label{appendix:formatting_generation}
 During inference, we can also leverage the special tokens $\texttt{\small{<|action|>}}$ and $\texttt{\small{<|observation|>}}$ as  ``beginning''- and ``end-of-subsequence'' markers to simplify the process of textual action decoding and generation. More precisely, LM agents trained on interaction history sequences with our method can be prompted to start generating an action prediction token-by-token by appending the special token $\texttt{\small{<|action|>}}$ to the end of a text prompt. Similarly, generation of the special token $\texttt{\small{<|observation|>}}$ can function as a termination condition for open-ended text action generation. Sampling actions from LM policies trained under our paradigm with these special tokens is thus easy to automate.

\newpage
\section{Environments}
\label{appendix:environments}

\subsection{BabyAI-Text}
\label{appendix:babyai}

As introduced in the main body of this paper, BabyAI-Text is a simple, procedurally generated grid-world environment with five constituent classes of task, introduced and open-sourced by \citet{carta2023grounding}. It is a fully text-based environment, reflecting the addition of a procedural, text wrapper over the well-known BabyAI environment developed by \citet{chevalier2018babyai}.

We provide a description of each of the task classes in BabyAI-Text below.
\begin{enumerate}
    \item \textbf{Go to \texttt{\small{<object>}}}: The agent must navigate to the item \texttt{\small{<object>}}.
    \item \textbf{Pick up \texttt{\small{<object>}}}: The agent must navigate to \texttt{\small{<object>}} and apply the ``pick up'' action.
    \item \textbf{Put \texttt{\small{<object A>}} next to \texttt{\small{<object B>}}}:  The agent must navigate to \texttt{\small{<object A>}} and pick it up. Then, it must navigate to \texttt{\small{<object B>}} and put down \texttt{\small{<object A>}}.
    \item \textbf{Open door}: The agent must navigate to the key with the same color as the target door and pick this key up. Then, it must navigate to the target door and applying the ``toggle'' unlocking action.
    \item \textbf{Pick up \texttt{\small{<object A>}} and then go to \texttt{\small{<object B>}}}:  The agent must navigate to \texttt{\small{<object A>}} and apply the ``pick up'' action before navigating to \texttt{\small{<object B>}}.
\end{enumerate}

\begin{figure}[h!]
    \centering
    \includegraphics[width=0.95\linewidth]{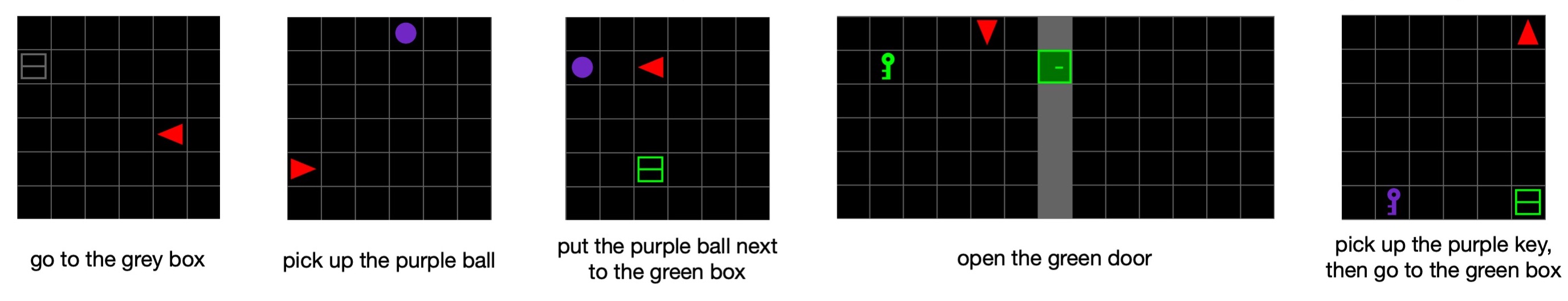}
    \caption{\textbf{BabyAI-Text}: Examples of visual observations for each of the five types of tasks in the BabyAI-Text environment, rendered in pixels. The agent is represented by the red triangle all images. The text-based description of the agent's task (as provided to neural LM agents in the ``instruction'' prefix of all tuning data and inference-time prompts) is provided below each image.}
    \label{fig:babyai_viz}
\end{figure}

A pixel-rendered visual observation of a task instance belonging to each of these classes is displayed in Figure \ref{fig:babyai_viz}. Across all tasks, there are five natural language actions available to the agent at each timestep: ``turn left,'' ``turn right,'' ``go forward,'' ``pick up,'' ``drop,'' ``toggle.'' We preserve the default tokenization of each of these actions in our neural LM agent experiments.

The reward assigned to a rollout across $t$ steps of interaction in BabyAI-Text is computed as
\begin{equation}
    r(t) = \mathbbm{1}(\text{task successfully completed at timestep } t) \cdot \left(1 - \frac{0.9 \cdot t}{t_{\max}}\right),
\end{equation}

where $t_{\max} \leq 200$ is a task-specific parameter and $\mathbbm{1}(\cdot)$ represents an indicator function. If an agent fails to complete a task in time $t = t_{\max}$, the rollout is assigned a reward of zero. We employ mean rollout reward over a batch of withheld, procedurally generated task instances as central our metric of neural LM agent generalization within this environment.

Examples of formatted full-text and \method{} interaction demonstrations collected in BabyAI-Text from the BabyAI bot \cite{chevalier2018babyai} via the procedures detailed in Section \ref{section:exp_data} are provided in Appendix \ref{appendix:babyai_history_examples}.

\subsection{NetHack}
\label{appendix:nethack}
NetHack is an exceptionally challenging and procedurally generated roguelike video game that was released in 1987. The game is rendered with ASCII-characters. Each instance or \textit{random seed} of NetHack features randomly re-generated player roles and attributes as well as seeded ``dungeon levels,'' special items, monsters, and much more. NetHack was also one of the very first games on the Web, uploaded by its creators to USENET newsgroups. As a result, human player data for NetHack is \textit{plentiful} \cite{craddock2021dungeon}. The symbolic NetHack bot \texttt{\small{AutoAscend}} continues to outperform all other decision-making agents for NetHack. The game is very far from solved \cite{hambro2022insights, hambro2022dungeons, piterbarg2023nethack, klissarov2023motif, zhong2022improving, tuyls2023scaling}. 

\subsubsection{Why is NetHack so hard?}
To win the game, a player must navigate through dozens of interlinked dungeon levels to the very ``bottom'' of the game world in order to retrieve a special item: the Amulet of Yendor. The player must then navigate all the way back up to the starting level, from which they must ``escape'' to conclude the game. This goal is dubbed as \textit{ascension} \cite{raymond2003guide}. It is exceedingly difficult for human expert players to achieve. Indeed, as of 01/30/2024, public play statistics on the NetHack altorg server\footnote{\texttt{https://alt.org/nethack/}} indicate that only only $\approx 0.96\% $ of all games in January 2024 have concluded in ascension.

Simply surviving in NetHack is also difficult. Throughout the game, monsters will randomly appear on the player's dungeon level. These monsters must be escaped from or vanquished for the game to go on. As the player descends deeper into the dungeon, monsters become increasingly harder to defeat, requiring mastery of increasingly complex weapons, fighting skills, and more. But monsters are not the only threats to a player's existence: the game may also end through ``death-by-starvation.'' To avoid this fate, players must continually search dungeon levels for food that is safe to eat \cite{raymond2003guide}.

Formally, NetHack a high-dimensional, infinite-horizon, stochastic, and partially-observable Markov decision process (POMDP) \cite{kaelbling1998planning, steinkraus2004combining}.  A ``fog-of-war'' obscures unvisited areas of dungeon levels, and the exact consequences of player actions may be randomized by the game engine. Furthermore, unlike many other games commonly used to evaluate neural agents such as MineCraft, the ``rule-set'' of NetHack is exceptionally difficult to accurately write down, let alone summarize and fit into a neural LM context length \cite{raymond2003guide, wu2023spring}. 

\subsubsection{NetHack Learning Environment}

\begin{figure}[h!]
    \centering
    \includegraphics[width=0.7\linewidth]{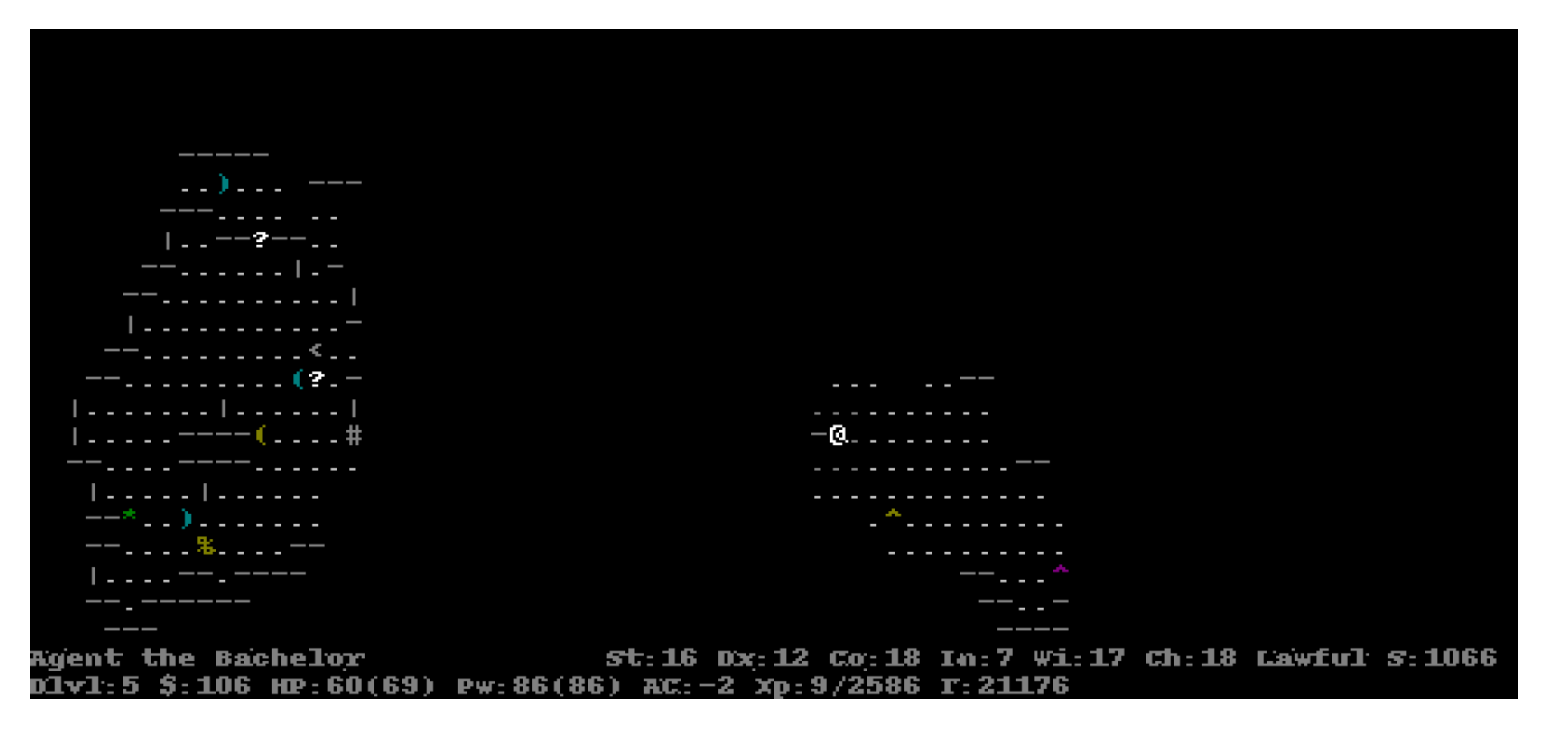}
    \caption{\textbf{NetHack Learning Environment}: An example of a visual observation in the NetHack Learning Environment, rendered in pixels. The agent is represented by the white ``\texttt{@}'' character in the middle of the image. The current state of the game is summarized at the bottom of the image, via ``bottom-line statistics.'' The agent is located on level 5 of the dungeon, as indicated by the text reading ``\texttt{Dlvl 5}'' in the bottom, left-hand corner.}
    \label{fig:nethack_viz}
\end{figure}

The NetHack Learning Environment (NLE) from \citet{kuttler2020nethack} wraps the NetHack video game ``as is'' into an environment for sequential decision-making research. NetHack \textit{game score} provides a measure of meaningful progress in the game, and is exposed by NLE to agents. This is the metric we use to evaluate neural agent performance and in-game generalization on NetHack in this paper. 

The full, default observation space of the environment has ten components, enumerated below.
\begin{enumerate}
    \item \textbf{\texttt{\small{glyphs}}}: A symbolic representation of all entities in-view to the player on the current dungeon-level.
    \item \textbf{\texttt{\small{message}}}: A text string representing the message from the game to the player (if any) at the current timestep of interaction.
    \item \textbf{\texttt{\small{blstats}}}: The ``bottom-line statistics'' of the game at the current timestep, which include an identifier for the current dungeon level, the total in-game score received, the player's hunger level, etc. 
    \item \textbf{\texttt{\small{tty\_chars}}}: The ASCII-character ``rendered'' version of the player's current observation. 
    \item \textbf{\texttt{\small{tty\_colors}}}: The color codes corresponding to each ASCII character in ``\texttt{\small{tty\_chars}}.''
    \item \textbf{\texttt{\small{tty\_cursor}}}:  The current location of the player's cursor over the ASCII-character view of the game.
    \item \textbf{\texttt{\small{inv\_glyphs}}}: A symbolic representation of all items currently in the player's inventory.
    \item \textbf{\texttt{\small{inv\_strs}}}: A byte array describing current inventory items.
    \item \textbf{\texttt{\small{inv\_letters}}}: The text character codes corresponding to each inventory item.
    \item \textbf{\texttt{\small{inv\_oclasses}}}: A symbolic representation of the \textit{class} of each inventory item.
\end{enumerate}
Most existing neural agents for NetHack directly leverage this observation space during learning. Indeed, prior to the writing of this paper, all state-of-the-art neural architectures for NetHack render the three \texttt{\small{tty*}} components\footnote{Note that, together, the three \texttt{\small{tty*}} components of the NLE observation space are equivalent to a human-player view of NetHack.} of the observation space in pixels (see Figure \ref{fig:nethack_viz}). There is a data writer native to NetHack that writes the \texttt{\small{tty*}} components of observations across timesteps of game-play to a custom \texttt{\small{.ttyrec}} format \cite{kuttler2020nethack}. Existing large scale datasets for NetHack store data in this format \cite{hambro2022dungeons, piterbarg2023nethack}.

There are several task variants of NLE. All of our experiments and investigations with NetHack focus on the hardest of these variants: the NetHack Challenge task, which exposes the full space of 121 possible per-timestep game actions to agents. This task was introduced during the NeurIPS 2021 NetHack Challenge Competition, which pitted symbolic decision-making agents against learned ones \cite{hambro2022insights}

As of 2023, there exists a natural language wrapper for NLE \cite{goodger2023nethack}, which translates the non-natural language components of NLE observations into equivalent descriptions across all tasks. This language wrapper also translates NLE actions into natural language equivalents and vice-versa. In the case of the ``NetHack Challenge'' NLE task that we study here, this corresponds to a per-timestep action space of 121 possible \textit{action strings}. A list of all possible commands (some of which correspond to \textit{sequences} of per-timestep actions) is given in the public NetHack wiki\footnote{\texttt{https://nethackwiki.com/wiki/Commands}}. We provide several annotated examples of NetHack actions below to give readers a feel for the game.

\begin{itemize}
    \item \textbf{\texttt{\small{northeast}}}: Move diagonally one tile in the northeast direction.
     \item \textbf{\texttt{\small{eat}}}: Eat the currently selected item (if any).
     \item \textbf{\texttt{\small{pickup}}}: Pickup the item under you (if any).
     \item \textbf{\texttt{\small{search}}}: Search the tiles immediately around you. This action may reveal passages or doors on the current dungeon level previously obscured from the player.
     \item \textbf{\texttt{\small{quaff}}}: Drink from the item under you or currently selected (if any).
     \item \textbf{\texttt{\small{loot}}}: Loot a box on the floor.
     \item \textbf{\texttt{\small{enhance}}}: Open an auxilary menu to check your current weapon skills.
\end{itemize}

Unfortunately, as stated in the main body of the paper, the NetHack language wrapper requires access to full observations, i.e. symbolic components of the NLE observation space \cite{goodger2023nethack}. We generate a new dataset with these observation components to support the study of text-based decision-making on NetHack in this work.

\newpage
\section{Dataset Statistics}
We provide data statistics reflecting all of the low-sample interaction demonstration datasets that we generate in this paper. A detailed description of the data generation procedures is provided in Section \ref{section:exp_data} of the main body. All token statistics are computed with the GPT-2 tokenizer, which is used for all LM experiments \cite{radford2019language}. Reported errors reflect one standard deviation across the full datasets.
\label{appendix:datasets}
\subsection{BabyAI-Text}
\label{appendix:babyai_dataset}

\begin{table*}[h!]
\label{table:babyai_data_stats}
\caption{\textbf{Processing BabyAI-Text 1K and 5K interaction history demonstration datasets}: Aggregate statistics reflecting the \method{} and full-text interaction history data formats that are used to tune LMs in our low-sample BabyAI-Text experiments. }
\begin{center}
\begin{small}
\begin{sc}
 \begin{tabular}{ lcccc }
 \toprule
      & \multicolumn{2}{c}{\texttt{BabyAI-Text-1K}} & \multicolumn{2}{c}{\texttt{BabyAI-Text-5K}} \\
       \cmidrule(r){2-3}\cmidrule(r){4-5} & \diff{} & Full-Text & \diff{} & Full-Text \\
 \midrule
 
 \abovespace
 \belowspace  History Horizon per Demo & 4 & 4 & 4 & 4 \\ \hline
 \abovespace Min Tokens Per Demo & 93 & 93 & 75 & 82 \\
 Max Tokens Per Demo & 667 & 392 & 667 & 392\\
 Median Tokens Per Demo & 358 & 232 & 354 & 230 \\
 \belowspace \textbf{Mean Tokens Per Demo} & 359 $\pm$ 103 & 231 $\pm$ 53 & \textbf{358 $\pm$ 99} & \textbf{230 $\pm$ 50} \\\hline
 \abovespace Min Tokens Per Obs & 21 & 12  & 21  & 12\\
 Max Tokens Per Obs & 266 & 122  & \textbf{266} & \textbf{123 } \\
 Median Tokens Per Obs & 130 & 57 & 128 & 57 \\
 \belowspace \textbf{Mean Tokens Per Obs} & 126 $\pm$ 51 & 59 $\pm$ 23 & \textbf{126 $\pm$ 49} & \textbf{59 $\pm$ 23} \\\hline
 \abovespace Min Tokens Per Action & 2 & 2 & 2 & 2 \\ 
 \belowspace Max Tokens Per Action & 3 & 3 & 3 & 3 \\  \hline  \hline 
   \abovespace\belowspace \textbf{Total Tokens} & \textbf{359K } &  \textbf{231K }& \textbf{1.78M} & \textbf{1.15M }\\
 \bottomrule
 \end{tabular}
 \end{sc}
 \end{small}
 \end{center}
 \vskip -0.1in
 \label{tab:agg_babyai_data_stats}
 \end{table*}

\subsection{NetHack}
\label{appendix:langhack}

\begin{table*}[h!]
\label{table:langhack_stats}
\caption{\textbf{Processing \dataset{} and \texttt{LangHack-2x} interaction history demonstration data}: Aggregate statistics reflecting all interaction history data formats that are used to train and tune models in our NetHack experiments.}
\begin{center}
\begin{small}
\begin{sc}
 \begin{tabular}{ lcccc }
 \toprule
      & \multicolumn{3}{c}{\dataset{} (500K)} & \multicolumn{1}{c}{\texttt{LangHack-2x} (1M)} \\
       \cmidrule(r){2-4}\cmidrule(r){5-5} & \diff{} & Full-Text & Pixel-Vector  &  \diff{}\\
 \midrule
 
 \abovespace
 \belowspace History Horizon Per Demo & 64 & 64 & 64 & 64 \\\hline
 \abovespace
 Min Tokens Per Demo & 544 & 13,785 & - & 544\\
 Max Tokens Per Demo & 54,887 & 105,301 & - & 73,469\\
 Median Tokens Per Demo & 7,915 &  35,508 & - & 7,911\\
 \belowspace \textbf{Mean Tokens Per Demo} & \textbf{8,465 $\pm$ 2,769} & \textbf{36,819 $\pm$ 9,557} & -& \textbf{8,464 $\pm $ 2,764}\\
 \hline \abovespace Min Tokens Per Obs & 1 & 196 & - & 1 \\
 \textbf{Max Tokens Per Obs} & \textbf{2269} & \textbf{1833} & - & \textbf{2269} \\
 Median Tokens Per Obs & 93 & 550 & - & 93 \\
 \belowspace \textbf{Mean Tokens Per Obs} & \textbf{123 $\pm$ 115} & \textbf{573 $\pm$ 156} & - &  \textbf{123 $\pm$ 115} \\ \hline
 \abovespace Min Tokens Per Action & 2 & 2 & - & 2\\ 
 \belowspace Max Tokens Per Action & 5 & 5 & - & 5\\
 \hline \hline
 \abovespace\belowspace
 \textbf{Total Tokens}  & \textbf{4.23B} & \textbf{18.4B} & - & \textbf{8.46B}\\
 \bottomrule
 \end{tabular}
 \end{sc}
 \end{small}
 \end{center}
 \vskip -0.1in
 \label{tab:agg_data_stats}
 \end{table*}

\newpage
\section{Examples of Formatted Interaction Histories}
\label{appendix:interaction_history_examples}
We provide a side-by-side comparison of example formatted full-text and \diff{} interaction histories in our two environments. These examples are not cherry-picked. For NetHack, we show only two consecutive timesteps of interaction history here in the interest of brevity.
\subsection{BabyAI-Text}
\label{appendix:babyai_history_examples}
\subsubsection{Instruction}
\textbf{Your task is to go to the yellow key.}\\
You can take 6 different actions: turn left, turn right, go forward, pick up, drop, and toggle.\\
a wall 4 steps forward\\
a wall 3 steps left
\begin{multicols}{2}
\subsubsection{Full-Text History Completion}
\texttt{<|action|>}\textbf{turn right}\\
\texttt{<|observation|>}\\
a wall 4 steps forward\\
a wall 3 steps right\\
\texttt{<|action|>}\textbf{turn right}\\
\texttt{<|observation|>}\\
a wall 3 steps forward\\
a wall 3 steps right\\
a yellow key 2 steps right and 2 steps forward\\
\texttt{<|action|>}\textbf{go forward}\\
\texttt{<|observation|>}\\
a wall 2 steps forward\\
a wall 3 steps right\\
a yellow key 2 steps right and 1 step forward\\
\texttt{<|action|>}\textbf{go forward}\\
\texttt{<|observation|>}\\
a wall 1 step forward\\
a yellow key 2 steps right\\
\texttt{<|action|>}\textbf{turn right}\\
\texttt{<|observation|>}\\
a wall 1 step left\\
a yellow key 2 steps forward\\
\texttt{<|action|>}\textbf{go forward}\\
\columnbreak

\subsubsection{\texttt{Diff} History Completion}
\texttt{<|action|>}\textbf{turn right}\\
\texttt{<|observation|>}\\
\texttt{@@} \texttt{-}2 \texttt{+}2 \texttt{@@}\\
\texttt{-}a wall 3 steps left\\
\texttt{+}a wall 3 steps right\\
\texttt{<|action|>}\textbf{turn right}\\
\texttt{<|observation|>}\\
\texttt{@@} \texttt{-}1 \texttt{+}1 \texttt{@@}\\
\texttt{-}a wall 4 steps forward\\
\texttt{+}a wall 3 steps forward\\
\texttt{@@} \texttt{-}2,0 \texttt{+}3 \texttt{@@}\\
\texttt{+}a yellow key 2 steps right and 2 steps forward\\
\texttt{<|action|>}\textbf{go forward}\\
\texttt{<|observation|>}\\
\texttt{@@} \texttt{-}1 \texttt{+}1 \texttt{@@}\\
\texttt{-}a wall 3 steps forward\\
\texttt{+}a wall 2 steps forward\\
\texttt{@@} \texttt{-}3 \texttt{+}3 \texttt{@@}\\
\texttt{-}a yellow key 2 steps right and 2 steps forward\\
\texttt{+}a yellow key 2 steps right and 1 step forward\\
\texttt{<|action|>}\textbf{go forward}\\
\texttt{<|observation|>}\\
\texttt{@@} \texttt{-}1,2 \texttt{+}1,1 \texttt{@@}\\
\texttt{-}a wall 2 steps forward\\
\texttt{-}a wall 3 steps right\\
\texttt{+}a wall 1 step forward\\
\texttt{@@} \texttt{-}1,3 \texttt{+}1,2 \texttt{@@}\\
\texttt{-}a yellow key 2 steps right and 1 step forward\\
\texttt{+}a yellow key 2 steps right\\
\texttt{<|action|>}\textbf{turn right}\\
\texttt{<|observation|>}\\
\texttt{@@} \texttt{-}1 \texttt{+}1 \texttt{@@}\\
\texttt{-}a wall 1 step forward\\
\texttt{+}a wall 1 step left\\
\texttt{@@} \texttt{-}1 \texttt{+}1 \texttt{@@}\\
\texttt{-}a yellow key 2 steps right\\
\texttt{+}a yellow key 2 steps forward\\
\texttt{<|action|>}\textbf{go forward}\\
\end{multicols}

\newpage

\subsection{NetHack}
\label{appendix:nethack_history_examples}
\subsubsection{Instruction}
\textbf{You are an agent playing NetHack. Predict the next keypresses.}\\
\texttt{statistics[}\\
Strength: 19/19\\
Dexterity: 14\\
Constitution: 18\\
Intelligence: 9\\
Wisdom: 7\\
Charisma: 8\\
Depth: 1\\
Gold: 25\\
HP: 57/61\\
Energy: 25/25\\
AC: 0\\
XP: 7/973\\
Time: 18818\\
Position: 43$|$13\\
Hunger: Not Hungry\\
Monster Level: 0\\
Encumbrance: Unencumbered\\
Dungeon Number: 0\\
Level Number: 1\\
Score: 3917\\
Alignment: Neutral\\
Condition: None\\
\texttt{]}\\
\texttt{glyphs[}\\
grave very far westsouthwest\\
boulder very far westnorthwest\\
orcish spear very far westnorthwest\\
stairs up very far westnorthwest\\
rocks very far westnorthwest\\
stairs down very far westnorthwest\\
iron skull caps very far westnorthwest\\
boulder far north, northnortheast, and northnorthwest\\
rock far northnortheast\\
rocks far north, westnorthwest, and northnorthwest\\
vertical wall far west\\
iron skull cap far westnorthwest and northnorthwest\\
hooded cloaks far northnorthwest\\
horizontal wall near north, south, southwest, and northwest\\
rock near northnortheast\\
boulder near westnorthwest\\
leather armor near northnorthwest\\
dark area very near southeast\\
tame large dog very near south\\
vertical wall adjacent northeast and east\\
rock adjacent southeast\\
hill orc corpse adjacent northwest\\
\texttt{]}\\
$\dots$
\begin{multicols}{2}
\subsubsection{Full-Text History Completion}
\tiny
\texttt{<|action|>}\textbf{northeast}\\
\texttt{<|observation|>}\\
\texttt{statistics}\\
Strength: 16/16\\
Dexterity: 12\\
Constitution: 18\\
Intelligence: 7\\
Wisdom: 17\\
Charisma: 18\\
Depth: 5\\
Gold: 106\\
HP: 55/69\\
Energy: 86/86\\
AC: -2\\
XP: 9/2586\\
Time: 21156\\
Position: 5$|$11\\
Hunger: Not Hungry\\
Monster Level: 0\\
Encumbrance: Unencumbered\\
Dungeon Number: 2\\
Level Number: 2\\
Score: 10660\\
Alignment: Lawful\\
Condition: None\\
]\\
\texttt{glyphs}\\
arrow far northnortheast\\
scroll labeled DAIYEN FOOELS far eastnortheast\\
stairs up far eastnortheast\\
key far east\\
scroll labeled KO BATE far east\\
flute far eastsoutheast\\
ape corpse far southsoutheast\\
horizontal wall far south\\
horizontal wall near northeast\\
southwest corner near southeast and southwest\\
arrow near southsoutheast\\
worthless piece of green glass near south\\
southeast corner adjacent north and west\\
northwest room corner adjacent northwest\\
]\\
$\dots$\\
\texttt{<|action|>}\textbf{search}\\
\texttt{<|observation|>}\\
\texttt{statistics}\\
Strength: 16/16\\
Dexterity: 12\\
Constitution: 18\\
Intelligence: 7\\
Wisdom: 17\\
Charisma: 18\\
Depth: 5\\
Gold: 106\\
HP: 55/69\\
Energy: 86/86\\
AC: -2\\
XP: 9/2586\\
Time: 21157\\
Position: 5$|$11\\
Hunger: Not Hungry\\
Monster Level: 0\\
Encumbrance: Unencumbered\\
Dungeon Number: 2\\
Level Number: 2\\
Score: 10660\\
Alignment: Lawful\\
Condition: None\\
]\\
\texttt{glyphs}\\
arrow far northnortheast\\
scroll labeled DAIYEN FOOELS far eastnortheast\\
stairs up far eastnortheast\\
key far east\\
scroll labeled KO BATE far east\\
flute far eastsoutheast\\
ape corpse far southsoutheast\\
horizontal wall far south\\
horizontal wall near northeast\\
southwest corner near southeast and southwest\\
arrow near southsoutheast\\
worthless piece of green glass near south\\
southeast corner adjacent north and west\\
northwest room corner adjacent northwest\\
]\\
$\dots$\\
\texttt{<|action|>}\textbf{northeast}
\\
$\dots$
\columnbreak

\subsubsection{\texttt{Diff} History Completion}
\normalsize
\texttt{<|action|>}\textbf{northeast}\\
\texttt{<|observation|>}\\
\texttt{<|@@|>} \texttt{-}15 \texttt{+}15 \texttt{<|@@|>}\\
\texttt{-}Position: 4$|$12\\
\texttt{+}Position: 5$|$11\\
\texttt{<|@@|>} \texttt{-}27 \texttt{+}26,0 \texttt{<|@@|>}\\
\texttt{-}horizontal wall far northeast\\
\texttt{<|@@|>} \texttt{-}30,3 \texttt{+}29,2 \texttt{<|@@|>}\\
\texttt{-}key far eastnortheast\\
\texttt{-}scroll labeled KO BATE far eastnortheast\\
\texttt{-}vertical wall far east\\
\texttt{+}key far east\\
\texttt{+}scroll labeled KO BATE far east\\
\texttt{<|@@|>} \texttt{-}34,9 \texttt{+}32,8 \texttt{<|@@|>}\\
\texttt{-}southeast corner far southeast\\
\texttt{-}arrow near southeast\\
\texttt{-}ape corpse near southeast\\
\texttt{-}worthless piece of green glass near southsoutheast\\
\texttt{-}northeast room corner near south\\
\texttt{-}southwest corner very near southwest\\
\texttt{-}vertical wall very near west\\
\texttt{-}southeast corner adjacent north\\
\texttt{-}horizontal wall adjacent northwest\\
\texttt{+}ape corpse far southsoutheast\\
\texttt{+}horizontal wall far south\\
\texttt{+}horizontal wall near northeast\\
\texttt{+}southwest corner near southeast and southwest\\
\texttt{+}arrow near southsoutheast\\
\texttt{+}worthless piece of green glass near south\\
\texttt{+}southeast corner adjacent north and west\\
\texttt{+}northwest room corner adjacent northwest\\
\texttt{<|action|>}\textbf{search}\\
\texttt{<|observation|>}\\
\texttt{<|@@|>} \texttt{-}14 \texttt{+}14 \texttt{<|@@|>}\\
\texttt{-}Time: 21156\\
\texttt{+}Time: 21157\\
\texttt{<|action|>}\textbf{northeast}\\
$\dots$

\end{multicols}

\newpage

\section{Tuning and Training Details}
\label{appendix:training_details}

\subsection{Neural Language Models}
All neural language agent experiments employ an off-the-shelf PyTorch \cite{paszke2017automatic} implementation of the smallest GPT-2 model \cite{radford2019language} distributed by the open-source HuggingFace Transformers library \cite{wolf2019huggingface}. This model has approximately 120 million parameters. Experiments with pretrained models employ the pretrained weights shipped with this distribution. Tokenization is similarly performed using the default GPT-2 tokenizer from HuggingFace. Our instruction finetuning procedures heavily draw from those open-sourced by \citet{wang2023far}.

\subsubsection{Extending GPT-2's Context Length}

As mentioned in the main body of the paper, we extend the context length of GPT-2 in our NetHack finetuning experiments by a factor of four from 1024 to 4096 tokens. This is done by adding new positional encodings to the model. The weights of new encodings are initialized from $\mathcal{N}(0,1)$, as is default in PyTorch.

\subsubsection{Hyperparameters}

We employ a batch size of 250 and a 32-epoch linear learning rate schedule with a warm-up ratio of $0.03$ in all training and finetuning experiments with the exception of the ultra-low data BabyAI-Text experiment (1K demonstrations only). In this experiment, we tune models with the same batch size and learning rate schedule but for 64 epochs. Note that when we first started conducting experiments, we did not expect such a larger number of consecutive finetuning epochs to be beneficial to LM agent performance. However, preliminary runs evaluating agent generalization with \textit{full rollouts} on held-out tasks confirmed this to be the case, with the effect particularly stark in our ultra-low sample BabyAI-Text finetuning settings.

Learning rate selection was performed once for full-text and \method{} experiments by sweeping over the set $\gamma \in \{1\cdot 10^{-3}, 3\cdot 10^{-4}, 5\cdot 10^{-5}, 1 \cdot 10^{-5}\}$. Models were tuned with action prediction losses for a single epoch on a $80/20$ split of each tuning dataset and were validated for perplexity on the held-out portion. We found $\gamma = 3 \cdot 10^{-4}$ to consistently be the best performing learning rate.

\subsubsection{Hardware}
All neural LM experiments were run using NVIDIA A100 GPU compute nodes with InfiniBand on an academic high performance computing cluster. BabyAI-Text finetuning experiments were carried out on single A100 GPUs, whereas NetHack finetuning experiments were conducted on two A100 GPUs at a time with a 48-hour budget per run. Model, optimizer, and data loader state checkpoints were saved frequently throughout the course of finetuning, and were used to restart and continue timed-out NetHack runs.

\subsubsection{ZeRO++}
Models were trained in mixed-precision with the \texttt{BFLOAT16} data format \cite{dean2012large} and in a distributed fashion using the Microsoft DeepSpeed and the HuggingFace Accelerate libraries \cite{accelerate}. We employ the highly performant ZeRO++ protocol for stage three sharding (i.e. sharding of optimizer states, gradients, and model parameters) together with model parameter CPU offloading to dramatically accelerate experiments \cite{ren2021zero, wang2023zero++}. We also use gradient accumulation to support training with the desired batch size ($B = 250$).

\subsection{NetHack Neural Vision Models}
\label{appendix:nethack_vision_model}
Vision-based state-of-the-art model architectures for NetHack (results reported as ``LSTM (CDPGT5)'' and ``Transformer'' in Table \ref{tab:agg_nethack_results}) are run exactly as originally implemented and open-soured during the NetHack Challenge Competition \cite{hambro2022insights} and by \citet{piterbarg2023nethack}, respectively. The Transformer-based model variant is referred to as a ``flat transformer'' in \citet{piterbarg2023nethack} and directly copies the architecture of CDPGT5, with the exception that the LSTM core module of the latter is replaced with a Transformer. 

The history length of all vision agents during training is initialized to the maximum history length of the \dataset{} dataset, $H = 64$. All other training hyperparameters of these models are set as in \citet{hambro2022dungeons} and \citet{piterbarg2023nethack}. Given our finding that very high epoch counts benefit the performance of tuned language agents, we similarly train vision agents for a very large number of epochs ($E = 32$) on \dataset{}. Model checkpoints are saved frequently during training. We employ three random seeds to test both model variants, randomizing weight initialization and data loading order in each experiment.

\subsubsection{On CDGPT5}

The LSTM-based CDPGT5 model architecture won third place in the neural track of the NeurIPS 2021 NetHack Challenge Competition, outperforming all other data-driven neural agents \cite{hambro2022insights}. It is also used ``off-the-shelf'' by \citet{tuyls2023scaling} in their imitation learning scaling experiments, where the authors set state-of-the-art performance for neural agents on NetHack by dramatically scaling up training time and sample count to train a large version of this model on 115B expert action labels from the leading NetHack bot \texttt{\small{AutoAscend}}. Despite these scaling efforts, their neural vision agent still significantly under-performs the expert \cite{tuyls2023scaling}.

\newpage
\section{Evaluation Details}
\label{appendix:eval_details}

\subsection{Neural Language Models}

\subsubsection{Philosophy}
Our goal in this paper is to improve the quality of LM generations in low-resource sequential decision-making. As a result, we minimize  the use of heuristics in our evaluations of neural language agents. Furthermore, rather than evaluating models by computing more standard language model metrics like perplexity over a validation dataset, we opt for a full ``rollout''-style evaluation. Though this choice makes our experiments much more costly, it
enables us to directly compare the rates of error accumulation associated with learned policies and adds rigour to our analysis.

At all timesteps of environment interaction, we generate token sequences from LMs with greedy decoding as implemented in HuggingFace Transformers \cite{wolf2019huggingface}. We impose no restrictions on generated tokens.

As alluded to in Appendix \ref{appendix:formatting_generation}, we prompt models to begin generating an action token subsequence by appending the special token \texttt{\small{<|action|>}} to the end of recent interaction history. We also implement a custom stopping criterion, halting generation as soon as the special token \texttt{\small{<|observation|>}} is decoded from an LM.

\subsubsection{Hardware}
All neural LMs are evaluated on single NVIDIA RTX8000, A100, or A4000 GPUs on an academic high performance computing cluster.

\subsubsection{BabyAI-Text}

We evaluate all neural language models on 256 withheld instances of the BabyAI-Text task variants used in tuning with interaction history horizon $h = H = 4$. The generation of incorrectly formatted or invalid action token sequences three times in a row results in rollout termination.

\subsubsection{NetHack}
As in our BabyAI-Text evaluation, all neural language agents tuned on NetHack demonstrations are evaluated in a standardized fashion. For each agent, we run a sweep over the maximal interaction history horizons $h \in \{4, 8, 15, 24, 64\}$ used to prompt LM action generation. In the event that $h$ timesteps of previous interaction history exceed maximal LM context lengths $(4096)$ at any given timestep, the history horizon included in the prompt is iteratively ``back-tracked'' across $h': 1\leq h' < h$ until this is no longer the case. The performance of models under each maximal history horizon is evaluated on a batch of 128 held-out instances of the NetHack video game via the NetHack Learning Environment (NLE) \cite{kuttler2020nethack}. 

We use precisely the same set of 128 game seeds in all evaluation sweeps. Furthermore, we employ the standard, ``NetHack Challenge'' variant of NLE in our evaluations, with inaction timeout set to $t = 100$ and no restrictions made over NetHack roles. 

The mean NetHack neural language agent game scores reported throughout the paper reflect the most performant interaction history horizon $h$. Standard errors are computed across the corresponding evaluation batch.

\subsection{NetHack Neural Vision Models}
NetHack neural vision agents are evaluated exactly like NetHack neural language agents, with one exception: faster action generation with these baselines enables us to use a larger evaluation batch size of 1024 withheld game instances and to run more extensive sweeps over pixel-vector interaction history horizon values provided to Transformer-based agents, $h \in \{1, 2, 4, 8, 15, 24, 32, 64\}$. Note that the hidden state of LSTM-based agents (i.e. models with the CDGPT5 architecture \cite{hambro2022insights}) are updated at each timestep throughout all games with no resetting, reflecting an effective agent interaction history horizon of $h = \infty$. The same evaluation procedure is conducted for checkpoints reflecting all $E = 32$ epochs of training. 

As is the case for our neural language agents score statistics, for all vision models, we report mean NetHack game score reflecting the most performant checkpoints and interaction history horizon (for Transformer-based models) in Figure \ref{fig:intro_figure} and Table \ref{tab:agg_nethack_results}.

Full learning curves of neural vision agent evaluation performance over the course of training (plotted to-scale against neural language model performance) are shown below in Figure \ref{fig:nethack_vision_learning_curves}.

\begin{figure}[h!]
     \centering
      \includegraphics[width=0.75\linewidth]{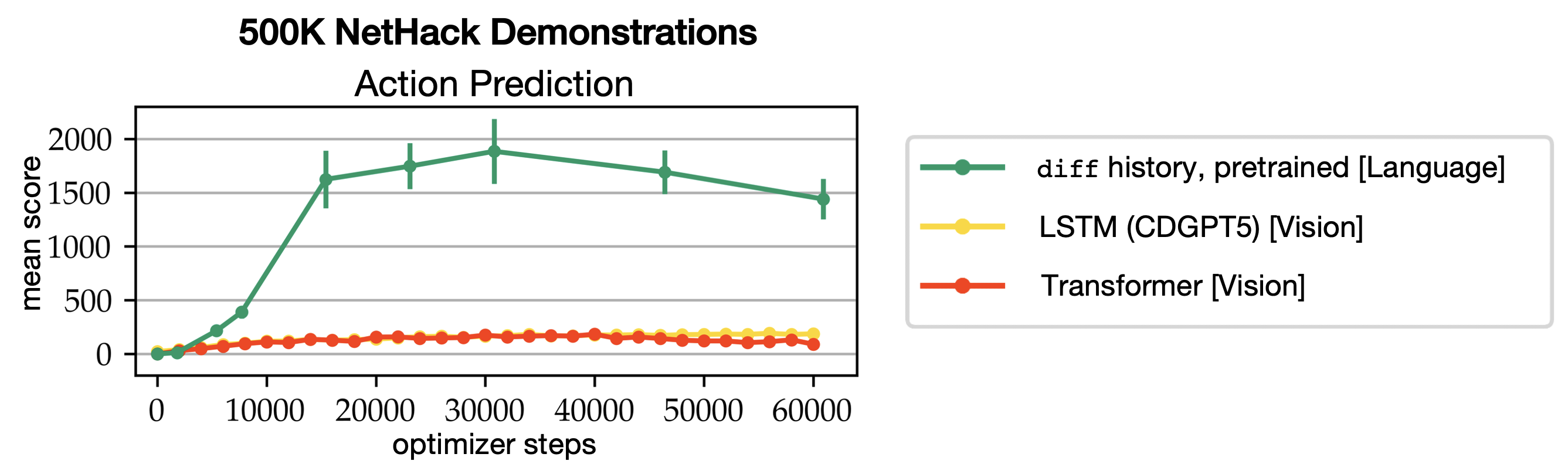}
      \label{fig:nethack_vision_learning_curves}
      \caption{\textbf{Training state-of-the-art vision baseline architectures on \dataset{}}: We provide learning curves reflecting the performance of neural vision agent baselines trained on \dataset{} vs \method{} neural language agents, to parallel those of Figure \ref{fig:nethack_learning_curves}. Error bars indicate one standard error over an evaluation batch of withheld game instances.}
      
      
    \end{figure}

\newpage
\section{ Scaling \texttt{diff} History in NetHack}
\label{appendix:scaling}
Below, we provide the learning curve of the \method{} scaling experiment described in Section \ref{section:results_scaling} as well as a visualization of the score distribution achieved by the scaled agent.

\begin{figure}[h!]
     \centering
      \includegraphics[width=0.6\linewidth]{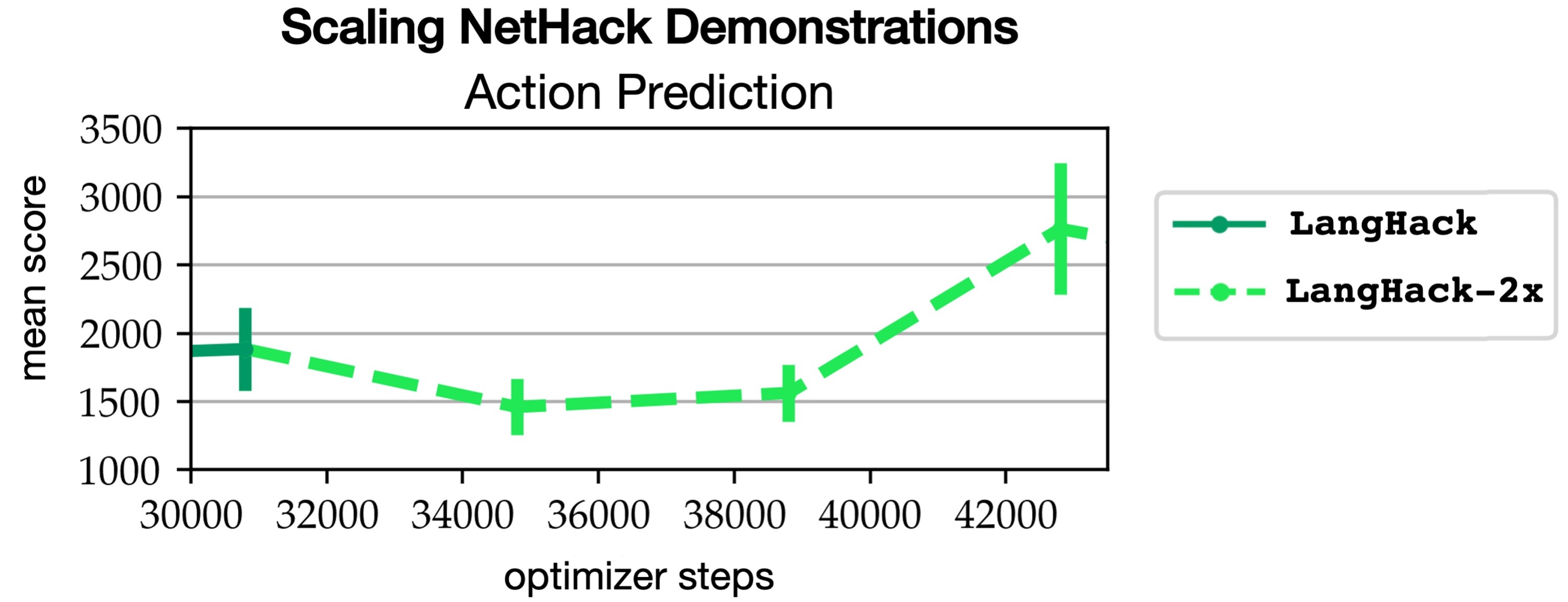}
      \caption{\textbf{Doubling NetHack interaction data}: We confirm that the performance of \method{} agents scales with data samples by tuning the best-performing \method{} agent checkpoint from our \dataset{} action prediction experiment on twice as many interaction demonstrations via \texttt{\small{LangHack-2x}}. This results in a 47\% gain in mean NetHack score (see Table \ref{tab:agg_nethack_results}). Error bars indicate one standard error in score over a batch of withheld games.}
      \label{fig:nethack_upscaled}
    \end{figure}

\vspace{1mm}
\begin{figure}[h!]
     \centering
      \includegraphics[width=0.45\linewidth]{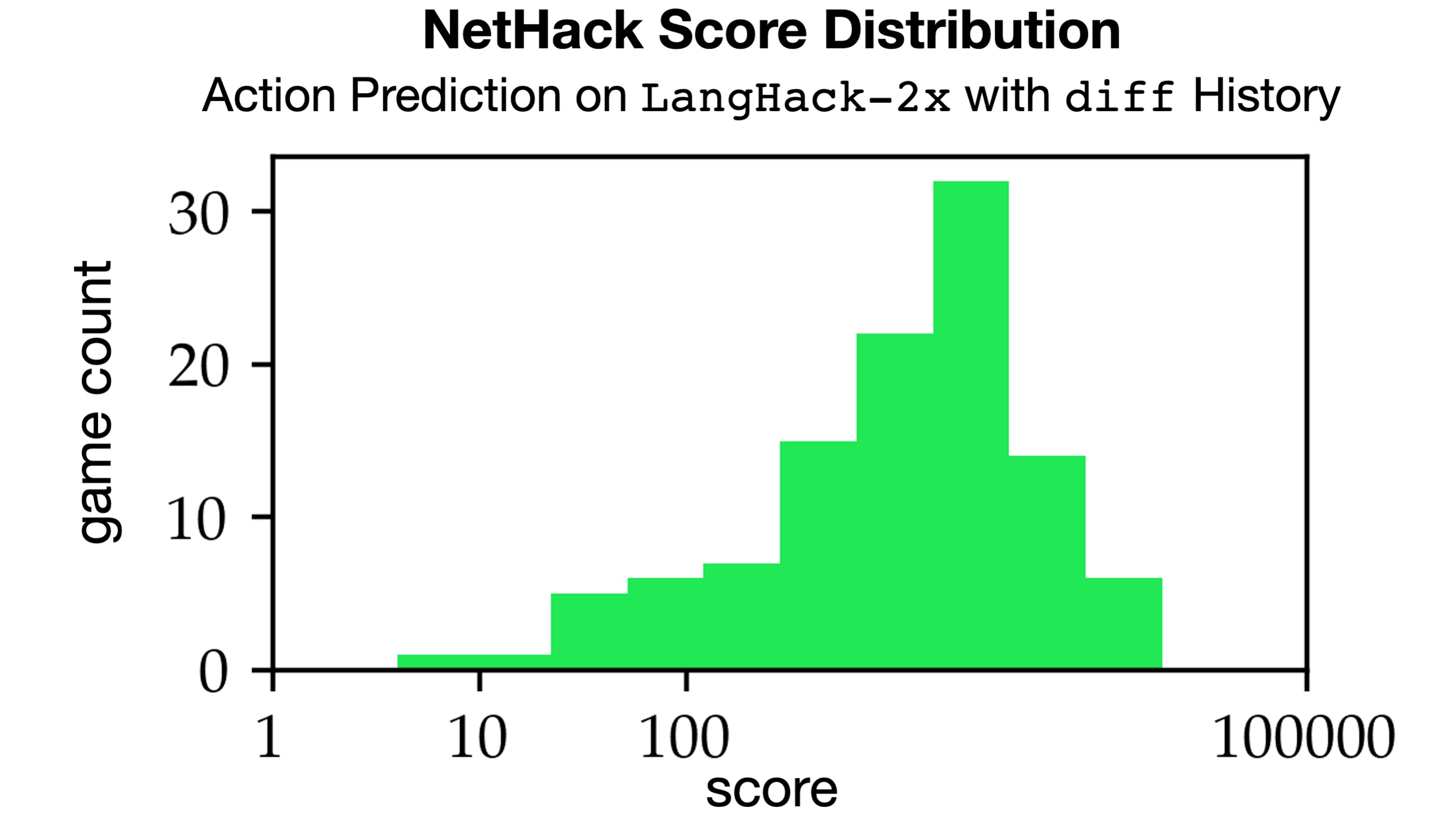}
      \caption{\textbf{NetHack game score distribution for best \texttt{LangHack-2x} checkpoint}: We demonstrate the distribution of game scores on withheld seeds of NetHack achieved by the most performant  \diff{} language agent trained on action prediction with \texttt{\small{LangHack-2x}} ($N=128$). We observe a long tail in the performance of the agent, with the LM achieving a score of $\approx20000$ and successfully descending down to level eight of the dungeon in one of the games. Similar phenomena have been reported in previous work \cite{hambro2022insights}. }
       \label{fig:nethack_upscaled_distribution}
    \end{figure}